\documentclass[10pt,journal,compsoc]{IEEEtran}
\usepackage[nocompress]{cite}

\usepackage[margin=0pt,font=small,labelfont=bf,labelsep=endash,tableposition=top]{caption}

\usepackage{amsmath}
\usepackage{array}

\usepackage[caption=false,font=footnotesize,labelfont=sf,textfont=sf]{subfig}

\usepackage{url}
\usepackage{graphicx}
\usepackage{amsmath}
\usepackage{amssymb}

\usepackage{epsfig}
\usepackage{epstopdf}
\usepackage{makecell,multirow,diagbox}
\usepackage{color}
\usepackage{soul}
\usepackage{url}
\usepackage{amsmath}
\usepackage{array}

\usepackage[utf8]{inputenc}
\usepackage{xcolor}
\usepackage[hidelinks]{hyperref}

\definecolor{yl_color}{RGB}{128, 255, 0}

\usepackage{mathtools,xspace}

\def\etal{{\it et al.}\xspace}

\hypersetup{
    colorlinks=true,
    breaklinks=false,
    urlcolor= blue,
    linkcolor= red,
    bookmarksopen=false,
    filecolor=black,
    citecolor=blue,
    linkbordercolor=blue
}
\newcommand{\methodName}{SPTS}

\begin{document}

\title{SPTS v2: Single-Point Scene Text Spotting}

\author{
Yuliang Liu, 
Jiaxin Zhang,
Dezhi Peng,
Mingxin Huang,
Xinyu Wang,
Jingqun Tang,
Can Huang, \\
Dahua Lin,
Chunhua Shen,
Xiang Bai, 
Lianwen Jin
\IEEEcompsocitemizethanks{
\IEEEcompsocthanksitem Y. Liu and X. Bai are with the School of Artificial Intelligence and Automation, Huazhong University of Science and Technology, Wuhan, 430074, China (email: \{ylliu, xbai\}@hust.edu.cn).
\IEEEcompsocthanksitem J. Zhang, J. Tang, and C. Huang are with the bytedance, Shanghai, China (email: zhangjiaxin.zjx1995@bytedance.com; tangjingqun@bytedance.com; can.huang@bytedance.com).
\IEEEcompsocthanksitem D. Peng, M. Huang, and L. Jin  are with South China University of Technology, Guangzhou, 510006, China (email: pengdzscut@foxmail.com; huangmingxin21@foxmail.com; eelwjin@scut.edu.cn). 
\IEEEcompsocthanksitem X. Wang and C. Shen are with Zhejiang University, Zhejiang, 310058, China (email: xinyu.wang02@adelaide.edu.au; chhshen@gmail.com). 
\IEEEcompsocthanksitem D. Lin is with the Chinese University of Hong Kong, 999077, China (email: dhlin@ie.cuhk.edu.hk). 
}
\thanks{Part of this work was done when Y. Liu was with the Chinese University of Hong Kong. Corresponding author: Lianwen Jin.}
}

\IEEEtitleabstractindextext{%
\begin{abstract}

    End-to-end scene text spotting has made significant progress due to its intrinsic synergy between text detection and recognition. Previous methods commonly regard manual annotations such as horizontal rectangles, rotated rectangles, quadrangles, and polygons as a prerequisite, which are much more expensive than using single-point. Our new framework, SPTS v2, allows us to train high-performing text-spotting models using a single-point annotation. SPTS v2 reserves the advantage of the auto-regressive Transformer with an Instance Assignment Decoder (IAD) through sequentially predicting the center points of all text instances inside the same predicting sequence, while with a Parallel Recognition Decoder (PRD) for text recognition in parallel, which significantly reduces the requirement of the length of the sequence. These two decoders share the same parameters and are interactively connected with a simple but effective information transmission process to pass the gradient and information. Comprehensive experiments on various existing benchmark datasets demonstrate the SPTS v2 can outperform previous state-of-the-art single-point text spotters with fewer parameters while achieving 19$\times$ faster inference speed. Within the context of our SPTS v2 framework, our experiments suggest a potential preference for single-point representation in scene text spotting when compared to other representations. Such an attempt provides a significant opportunity for scene text spotting applications beyond the realms of existing paradigms. Code is available at: \url{https://github.com/Yuliang-Liu/SPTSv2}.
\end{abstract}

}

\maketitle

%
%

\IEEEdisplaynontitleabstractindextext
\IEEEpeerreviewmaketitle

\IEEEraisesectionheading{\section{Introduction}
\label{sec:introduction}}
    \IEEEPARstart{S}{c}ene text reading techniques have made great strides in recent years. Given an image, text spotters can simultaneously locate and recognize the textual content, enabling many real-world applications such as document digitalization, intelligent assistants, and autopilot. Basically, bounding boxes such as rectangles, quadrilaterals, and polygons are commonly employed to represent the text of different shapes. However, the fact that humans can intuitively read texts without such a defined region encourages the development of a bounding-box-free text spotter, lifting the limitations imposed by bounding-box annotations.

    \begin{figure}[t!]
	\centering
	\includegraphics[width=0.8\linewidth]{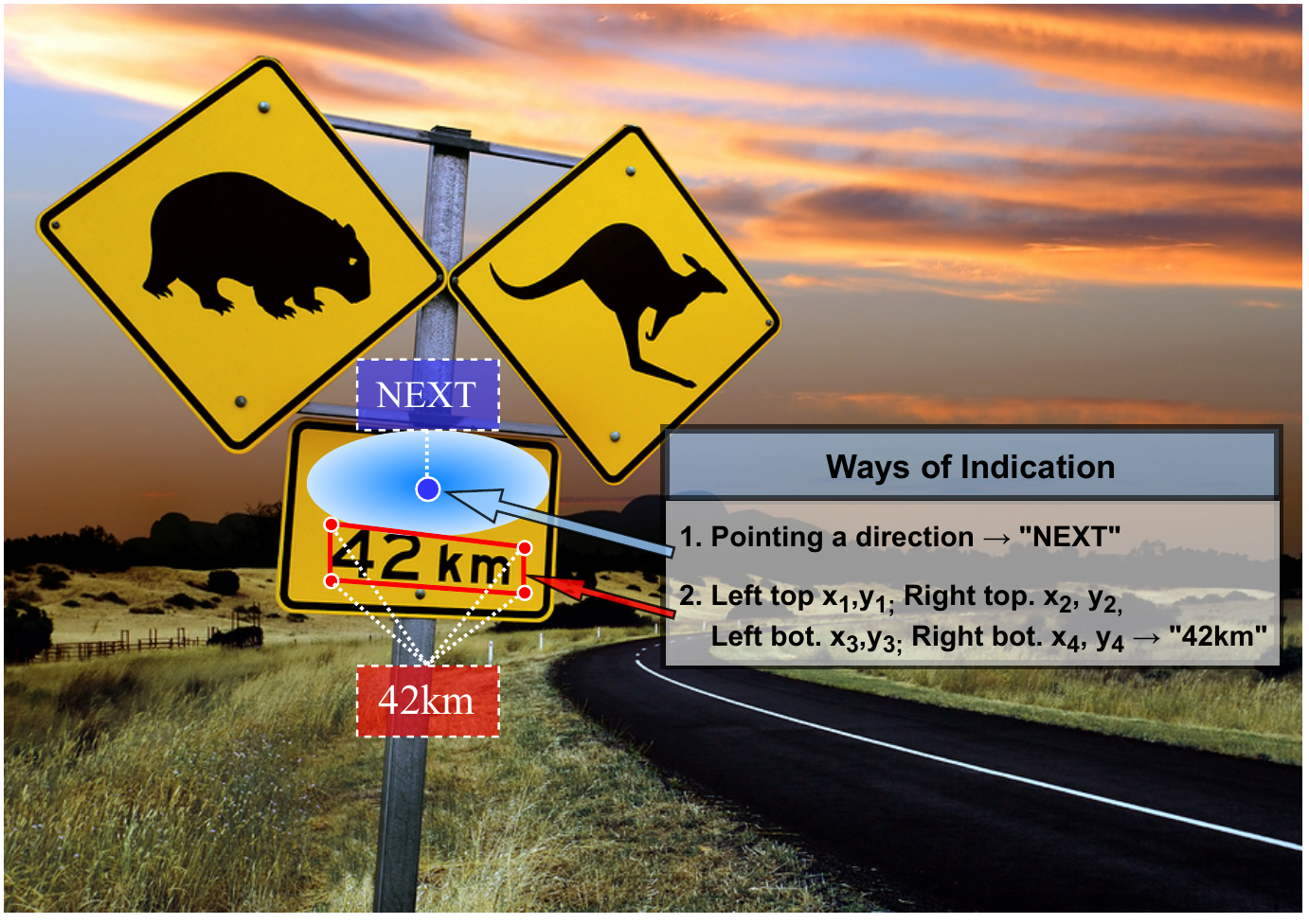}
	\caption{Existing OCR methods typically use bounding boxes to  represent the text area. However, inspired by how humans can intuitively read texts without such a defined region, this paper demonstrates that a single point is sufficient for guiding the model to learn a strong scene text spotter.}
	\label{fig:intro_indication}
    \end{figure}

    As shown in Fig.~\ref{fig:intro_indication}, the previous methods use a bounding box consisting of a series of coordinates to define the instance-level text, where the enclosed region is considered a positive sample. With its simplicity and straightforwardness, the bounding box has become the favored annotation format for many other vision tasks. However, unlike the target in object detection tasks that are usually presented in a defined appearance, text instances may appear in arbitrary shapes due to the different typographies and fonts. Therefore, it is required to use bounding boxes containing more coordinates, such as polygons, to label these arbitrarily shaped texts. Otherwise, considerable noise might be involved, which may negatively impact the recognition performance. For example, Total-Text~\cite{ch2017total} uses up to 20 coordinates while SCUT-CTW1500~\cite{liu2017deep} uses up to 28 coordinates to annotate a single curved scene text instance. Although using polygons can, to some extent, alleviate the problem of noise in labeling arbitrarily shaped text, it greatly increases the annotation cost. To solve these issues, this paper proposes a new manner of supervision for scene text spotters by using a single guidance point. As shown in Fig.~\ref{fig:intro_indication}, each of the texts is indicated by a single point within the instance. Such a streamlined representation breaks the limitation of bounding boxes, enabling the model to access the pixels in vicinity freely and further learn to discriminate the boundary between texts. Moreover, it considerably saves the cost of annotation compared to the polygons.

    \begin{figure*}[t!]
    \centering
    \begin{minipage}[c]{.19\linewidth}
      \includegraphics[width = 3.3cm, height = 3.2cm]{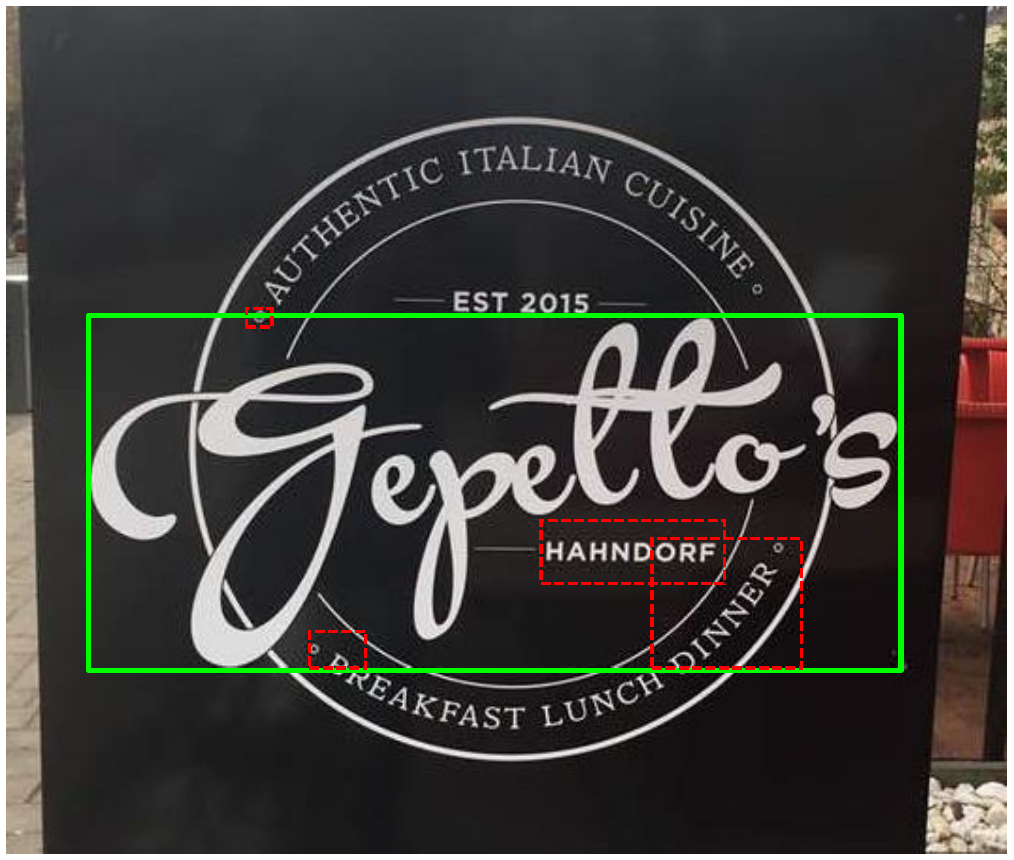}  
      \centerline{\small{(a) Rectangle (55s)}}
    \end{minipage}    
    \begin{minipage}[c]{.19\linewidth}
      \includegraphics[width = 3.3cm, height = 3.2cm]{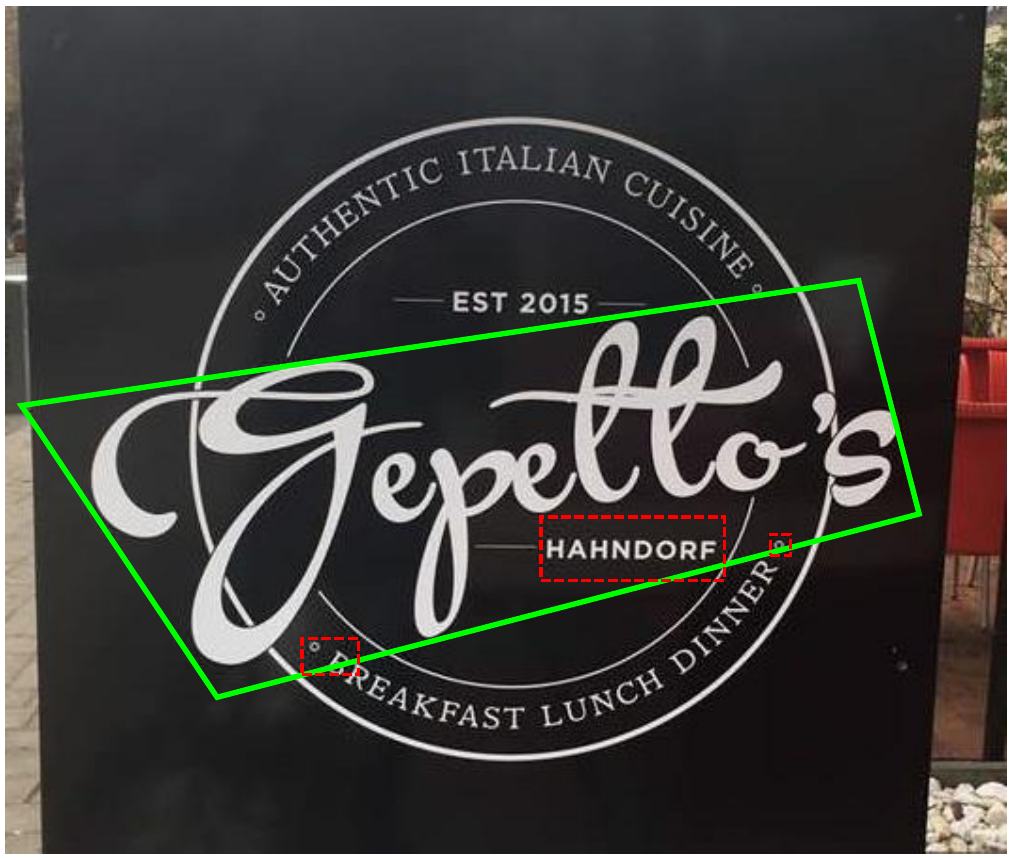}  
      \centerline{\small{(b) Quadrilateral (96s)}}
    \end{minipage}
    \begin{minipage}[c]{.19\linewidth}
      \includegraphics[width = 3.3cm, height = 3.2cm]{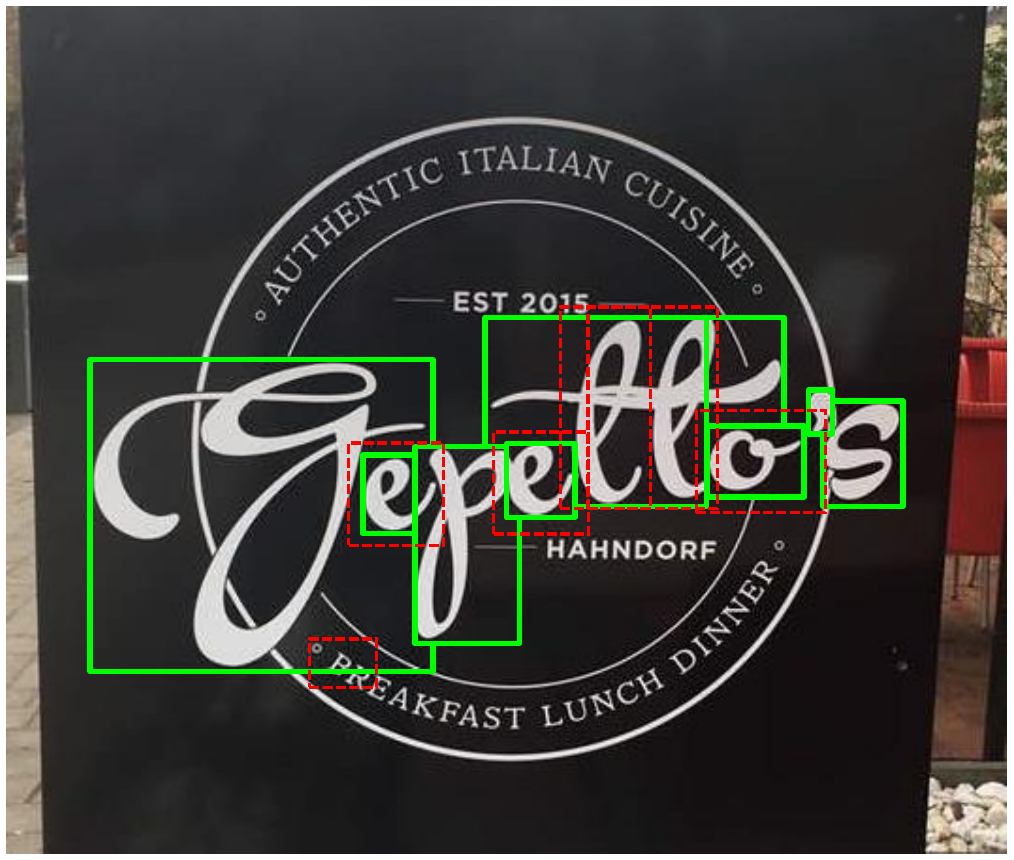}  
      \centerline{\small{(c) Character (581s)}}
    \end{minipage}
    \begin{minipage}[c]{.19\linewidth}
      \includegraphics[width = 3.3cm, height = 3.2cm]{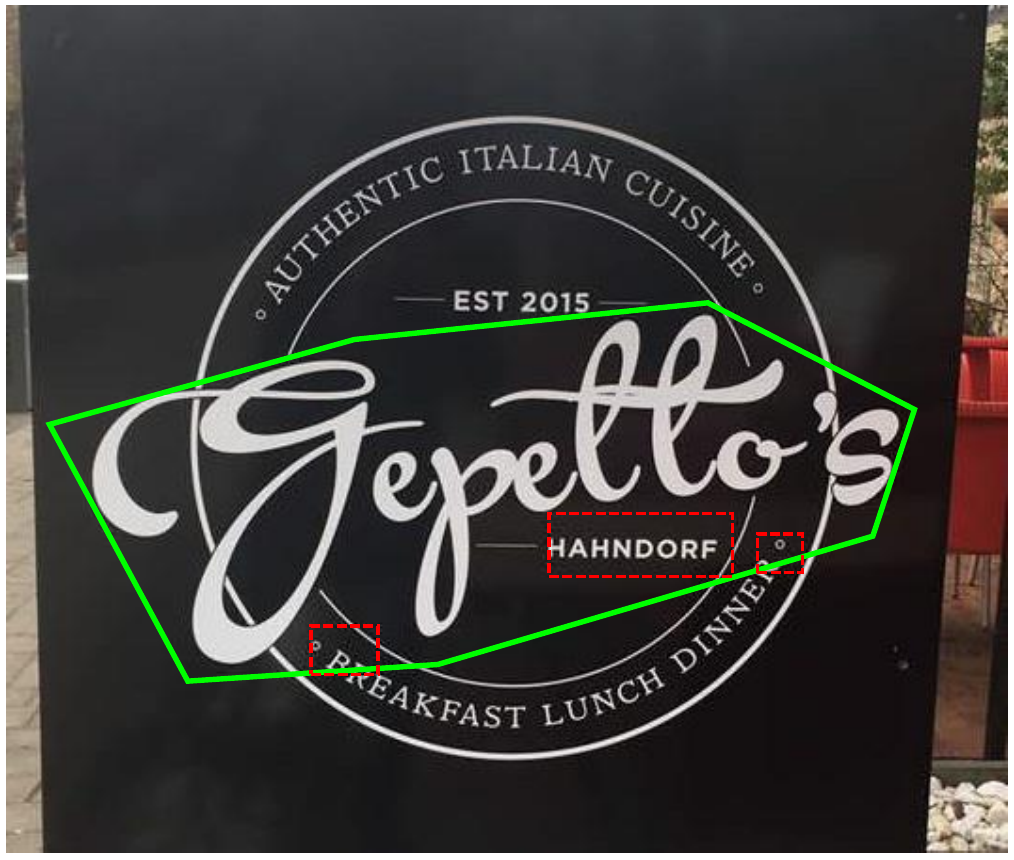}
      \centerline{\small{(d) Polygon (172s)}}
    \end{minipage}
    \begin{minipage}[c]{.19\linewidth}
      \includegraphics[width = 3.3cm, height = 3.2cm]{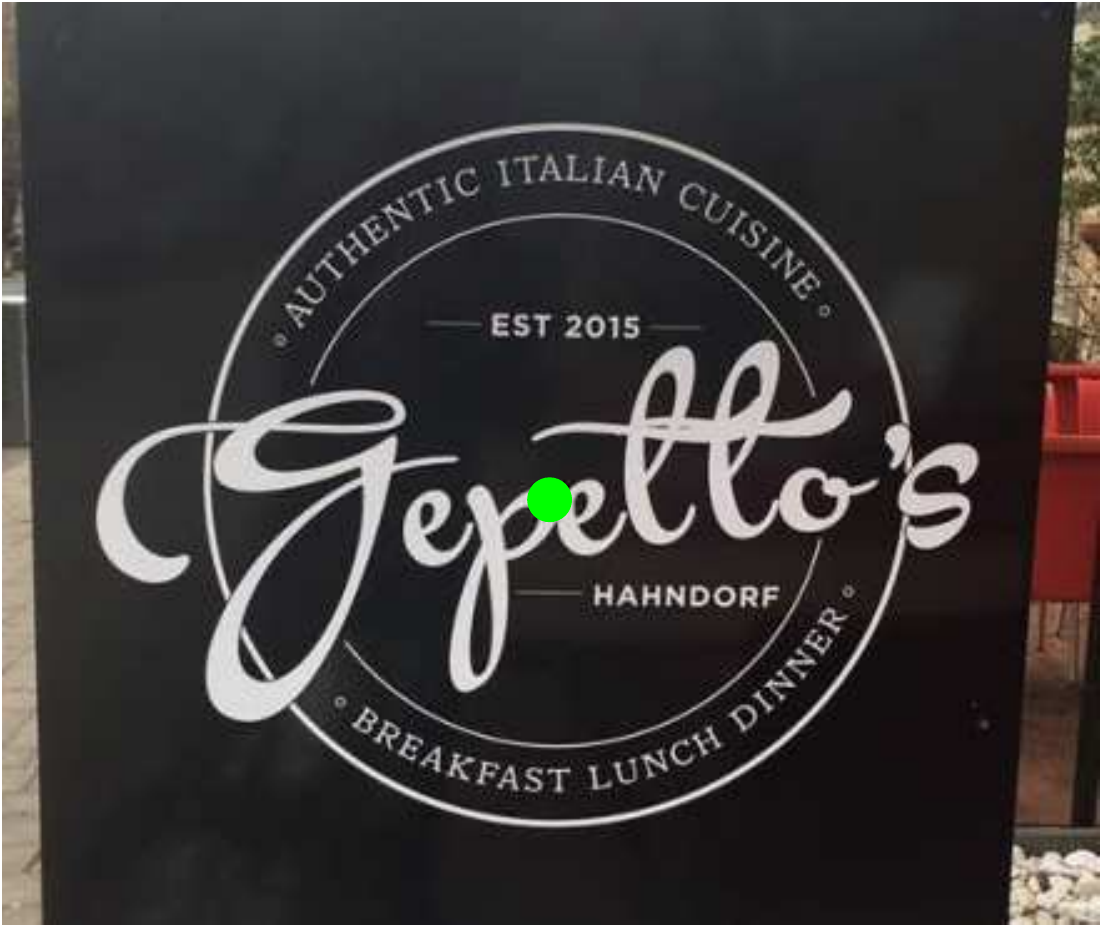}  
      \centerline{\small{(e) Single-Point (11s)}}
    \end{minipage}
    \caption{\textbf{Different annotation styles and their time cost} (for all the text instances in the sample image) measured by the LabelMe\protect\footnotemark tool. Green areas are positive samples, while red dashed boxes are noises that may be possibly included. Note the time is measured by the average of three annotators. For representations other than point, they normally need to zoom in for alignment of the exact location, which consumes non-trivial annotation efforts. }
    \label{fig:trn_tgt}
    \end{figure*}

\begin{figure}[t!]
    \centering
        \begin{minipage}{0.49\linewidth}
            \includegraphics[width=4.2cm, height=2.4cm]{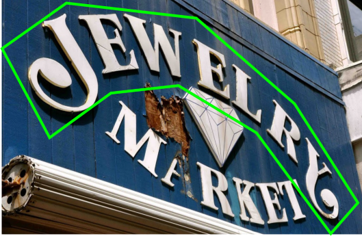}
            \centerline{\small{(a) Mask TextSpotter~\cite{liao2019mask}}}
            \label{fig:mts}
        \end{minipage}
        \begin{minipage}{0.49\linewidth}
            \includegraphics[width=4.2cm, height=2.4cm]{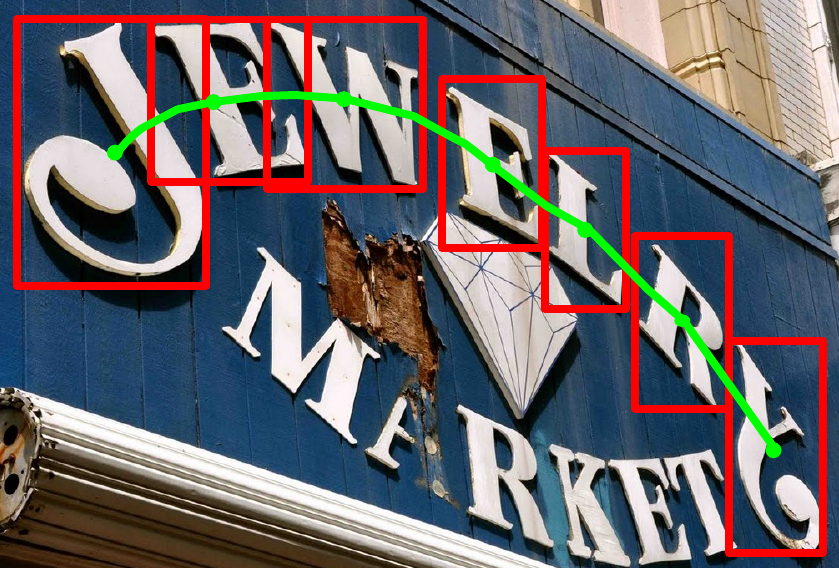}
            \centerline{\small{(b) Text Dragon~\cite{feng2019textdragon}}}
            \label{fig:textdragon}
        \end{minipage}
        \begin{minipage}{0.49\linewidth}
            \includegraphics[width=4.2cm, height=2.4cm]{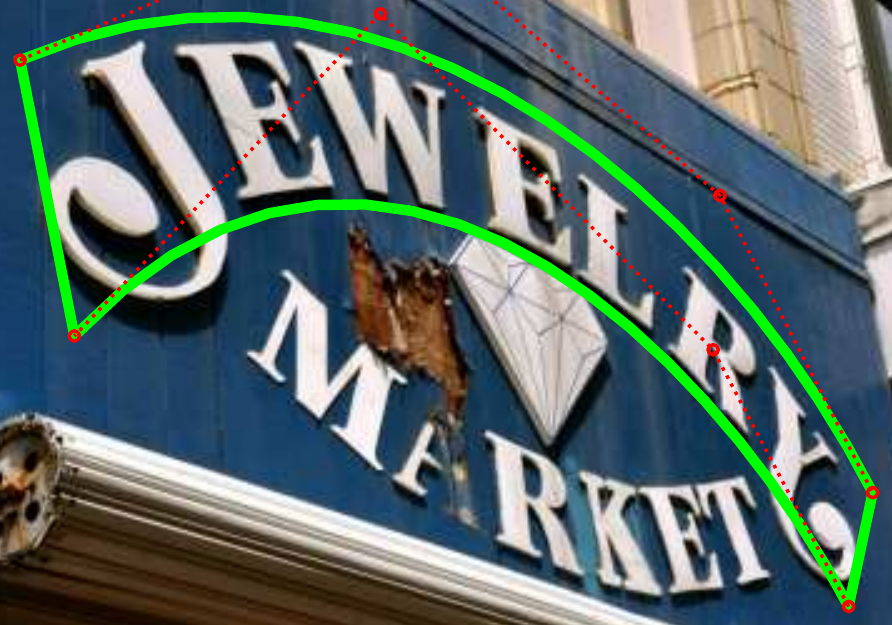}
            \centerline{\small{(c) ABCNet~\cite{liu2020abcnet}}}
            \label{fig:bzcurve}
        \end{minipage}
        \begin{minipage}{0.49\linewidth}
            \includegraphics[width=4.2cm, height=2.4cm]{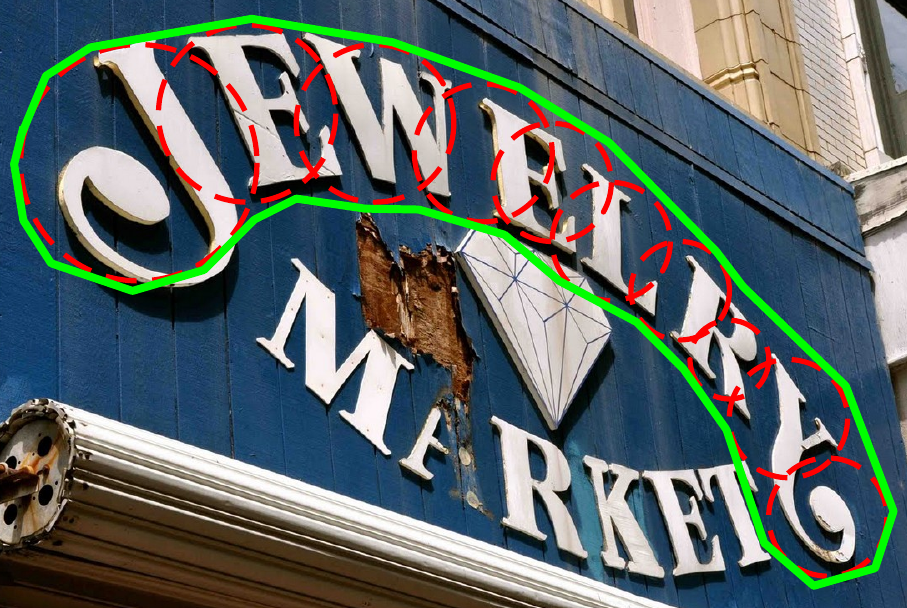}
            \centerline{\small{(d) Text Snake~\cite{long2018textsnake}}}
            \label{fig:textsnake}
        \end{minipage}
        \caption{Different representations of text instances. 
        }\label{fig:novel_representations}
    \end{figure}

    In the past decade, the focus of research on scene text spotting has shifted from horizontal~\cite{tian2016detecting, liao2017textboxes} and multi-oriented text~\cite{zhou2017east, yao2012detecting,liu2017deep} to arbitrarily shaped text~\cite{liu2020abcnet, lyu2018mask}, as reflected in the transition from rectangular and quadrilateral annotations to more compact but more expensive polygons. As shown in Fig.~\ref{fig:trn_tgt}, the rectangular bounding boxes are prone to involve other text instances, which may confuse the subsequent scene text recognition. Furthermore, many efforts have been made to develop more sophisticated representations to fit arbitrarily shaped text instances~\cite{shi2017detecting, feng2019textdragon, wang2020textray, liu2020abcnet, zhu2021fourier, long2018textsnake}. For example, as shown in Fig.~\ref{fig:novel_representations}, Mask TextSpotter~\cite{liao2019mask} utilizes boundary polygons to localize the text region. Text Dragon~\cite{feng2019textdragon} utilizes character-level bounding boxes to generate center-lines for enabling the prediction of local geometry attributes, ABCNet~\cite{liu2020abcnet} converts polygon annotations to Bezierd-curves for representing curved text instances, and Text Snake~\cite{long2018textsnake} describes text instances by a series of ordered disks centered at symmetric axes. 
    These heuristic representations are carefully designed by knowledgeable experts. Although they have been shown to be effective for aligning features between text detection and recognition modules, the reliance on manually-designed rules undeniably undermines the generalizability. Specifically, specified network architectures and modules are required to process the features and annotations, such as variants of RoI modules and post-processing mechanisms. In addition, as shown in Fig.~\ref{fig:trn_tgt}, the above representations that rely on annotations of polygonal or character bounding boxes are costly for labeling, while the proposed single point can halve the cost.

    In the past few years, some researchers~\cite{tian2017wetext, bartz2018see, hu2017wordsup, baek2019character} have explored training the OCR models with coarse annotations in a weakly-supervised manner. These methods can mainly be separated into two categories, \emph{i.e.}, (1) bootstrapping labels to finer granularity~\cite{tian2017wetext, bartz2018see} and (2) training with partial annotations~\cite{hu2017wordsup, baek2019character}. The former usually derives character-level labels from word-level or line-level annotations; thus, the models could enjoy the well-understood advantage of character-level supervision without introducing overhead costs. The latter is committed to achieving competitive performance with fewer training samples. However, both methods still rely on the bounding box annotations. A recent study~\cite{peng2022spts} demonstrates that point-only annotation for scene text can still achieve competitive performance in scene text spotting tasks.  

    One of the underlying problems to replace the bounding box with a simpler annotation format, such as a single-point, is that most text spotters rely on RoI-like sampling strategies to extract the shared backbone features. For example, Li \etal~\cite{li2017towards} and Mask TextSpotter require box and mask prediction inside an RoI~\cite{liao2020masktext}; ABCNet~\cite{liu2020abcnet} proposes BezierAlign to wrap the curved representation into the horizontal format, while TextDragon~\cite{feng2019textdragon} introduces RoISlide to unify the detection and recognition heads. 

    In this paper, inspired by the recent success of a sequence-based object detector Pix2Seq~\cite{chen2021pix2seq}, we show that the text spotter can be trained with a single point. Due to the concise form of annotation, annotation time can be significantly saved, \emph{e.g.}, it only takes less than one-fiftieth of the time to label single-points for the sample image shown in Fig.~\ref{fig:trn_tgt} than annotating character-level bounding boxes. Another motivating factor in selecting point annotation is that a clean and efficient OCR pipeline can be developed, discarding the complex post-processing module and roi-based sampling strategies; thus, the ambiguity introduced by RoIs (see red dashed regions in Fig.~\ref{fig:trn_tgt}) can be alleviated.

    However, adopting single-point representation is still challenging. The previous state-of-the-art single-point text spotter (SPTS) in our conference version~\cite{peng2022spts} uses the auto-regressive Transformer to generate the long sequence for all text instances. Here, point-only prediction is very straightforward and can handle unordered text instances as a whole. Thus it can learn to avoid repeated predictions, specified label assign methods like bipartite matching in DETR~\cite{carion2020end}, or complicated post-processing like non-maximum-suppression (NMS)~\cite{neubeck2006efficient}. Although our conference version SPTS~\cite{peng2022spts} is effective, the inference speed of it is extremely low, especially the images where many instances are simultaneously included. 

    Therefore, to take advantage of auto-regressive prediction and preserve high efficiency at the same time, \textit{we design SPTS v2, which significantly improves the inference speed while achieving higher performance.} Specifically, we observe that the long sequence of the result is mainly occupied by the text recognition, and thus we can only predict the location, \emph{e.g.}, $x$ and $y$ for each instance auto-regressively, termed as an indicator, in the first instance assignment decoder (IAD); while for the second parallel recognition decoder (PRD), each indicator is responsible for its subsequent text recognition prediction, which can be implemented in parallel. The rationale behind this is that the IAD is to solve implicit label assignment while PRD is for parallel prediction of the responsible text recognition result given the features after the label assignment. To enable the gradient flow from recognition features to the IAD stage, we propose a simple yet effective information transmission method that integrates the embeddings of the text location and the features in IAD stage, which is demonstrated essential to the success of the SPTS v2. 

    We summarize our contributions as follows:
    \begin{itemize}
        \item We form text spotting as a language modeling task, using only a cross-entropy loss. This streamlined approach eliminates the need for intricate post-processing and sampling strategies, offering increased flexibility.
        \item SPTS v2 introduces a new Instance Assignment Decoder (IAD) as well as the Parallel Recognition Decoder (PRD) by sharing the same parameters, which are incorporated by a simple but effective information transmission method. SPTS v2 significantly reduces the length of the sequence, outperforming the conference version SPTS, with fewer parameters and 19x faster inference speed. 
        \item Extensive experiments conducted on five benchmarks, \emph{i.e.}, ICDAR 2013~\cite{karatzas2013icdar}, ICDAR 2015~\cite{karatzas2015icdar}, Total-Text~\cite{ch2017total}, SCUT-CTW1500~\cite{liu2019curved}, and Inverse-Text~\cite{ye2022dptext} involving both horizontal and arbitrarily shaped texts demonstrate the competitive performance of our method against previous state-of-the-arts. 
    \end{itemize}
    \footnotetext{\sf  https://github.com/wkentaro/labelme}

    \section{Related Work}
    \label{sec:related-works}
    In the past decades, a variety of annotation styles have been proposed focusing on various scenarios of scene text spotting, including letters represented by stroke-level or character-level bounding box~\cite{xing2019convolutional,lyu2018mask,liao2019mask,liao2020masktext,qiao2021mango}, horizontal text~\cite{karatzas2013icdar, karatzas2011icdar} represented by rectangles (Fig.~\ref{fig:trn_tgt}(a)), multi-oriented text~\cite{karatzas2015icdar, nayef2019icdar2019} represented by quadrilaterals (Fig.~\ref{fig:trn_tgt}(b)), arbitrary-shaped text~\cite{ch2017total, liu2019curved, chng2019icdar2019} represented by polygons (Fig.~\ref{fig:trn_tgt}(d)), and other novel representation such as single-point~\cite{peng2022spts,fan2022pointly} or non-point~\cite{peng2022spts,tang2022you}. 

    \subsection{Character-level Scene Text Spotter}
    In the early stage, many classical methods require character-level annotations to train the model. Wang \etal~\cite{wang2010word} uses a character classifier based on the HOG~\cite{dalal2005histograms} features to read text. Bissacco \etal~\cite{bissacco2013photoocr} combine DNN with HOG features to build a character classifier system for text recognition. The follow-up work~\cite{jaderberg2014deep} further develops a convolutional neural network to be the character classifier. The above methods are adjusted to the horizontal text using character-level annotations. Some researchers attempt to extend the character-level scene text spotter to handle the arbitrarily-shaped text. Mask TextSpotter~\cite{lyu2018mask} designs a character segmentation module to locate and recognize the character; its improved version~\cite{liao2019mask, liao2020masktext} significantly reduce the cost of manual annotations. CharNet~\cite{xing2019convolutional} proposes a one-stage framework to boost the text spotting performance by utilizing character-level annotations. CFRATS~\cite{baek2020character} locates the character regions and sends the information of the character regions to the attention-based recognizer. The MANGO~\cite{qiao2021mango} develops a position-aware mask attention module to generate a location mask for the character and use a sequence decoder to obtain the recognition results.
    
    \subsection{Rectangle-based Scene Text Spotter}
    The rectangle-based scene text spotter plays an important role in the early stage of the task. Weinman \etal~\cite{weinman2013toward} propose a text spotting system that first generates the text proposal and then use an independent word recognition model to extract the text content. Li \etal~\cite{li2017towards} adapt a generic object detectors framework Faster R-CNN~\cite{ren2015faster} to detect the rectangle-shape text and bridge the detector and CTC-based~\cite{graves2006connectionist} recognizer through sharing backbone. Its enhanced version~\cite{wang2021towards}, equipped with a 2D attention recognition module, is designed to handle the irregular text. Gupta \etal~\cite{gupta2016synthetic} employ FCRN to detect the rectangle boxes and use a word classifier as the recognizer. Recently, Liao \etal~\cite{liao2018textboxes++} propose a text spotting systems based on the TextBoxes~\cite{liao2017textboxes} and the CRNN~\cite{shi2017end}, which are used to locate and recognize the words, respectively. Shi \etal~\cite{shi2018aster} use the TextBoxes as the detector to obtain the detection results and propose a new recognizer ASTER~\cite{bookstein1989principal} which develops a Thin-Plate Spline transformation as rectification network to rectify the recognition image.
    
    \subsection{Multi-oriented Scene Text Spotter}
    
    Recent methods develop multi-oriented scene text spotters to handle text instances with complex shapes. Busta \etal~\cite{busta2017deep} propose a Deep TextSpotter which uses the YOLOv2~\cite{redmon2017yolo9000} as the detector to detect the multi-oriented text and use a CTC-based recognizer to transform the recognition features into a sequence of characters. FOTS~\cite{liu2018fots} proposes a new RoI operation, termed RoI Rotate, to transform oriented text recognition features to regular ones from quadrilateral detection results. He \etal~\cite{he2018end} propose a similar framework to locate the instance of the text. They further develop a Text-Alignment to sample rotated features into the horizontal feature and use an attention-based recognizer to improve the performance.

    \subsection{Arbitrarily-shaped Scene Text Spotter}
    Arbitrarily-shaped scene text spotting is challenging due to the diversity of text, such as various shapes, colors, fonts, and languages, which has attracted increasing attention. In this aspect, the text is normally annotated with polygons of arbitrary shapes. Recently, Qin \etal~\cite{qin2019towards} propose an RoI Masking to suppress the background noise for the recognition features and use a 2D attention-based recognizer to read the arbitrarily-shaped text from recognition features. Wang \etal~\cite{wang2021pan++} design a spotting system, termed PAN++, based on the fast detector PAN~\cite{wang2019efficient}. TextNet~\cite{sun2018textnet} predicts the quadrilateral text proposal to locate the text and develops a perspective RoI transformation process to rectify the quadrilateral features. Feng \etal~\cite{feng2019textdragon} describe the text instances as a series of quadrangles and propose the RoISlide to connect the quadrangles for text recognition. Wang \etal~\cite{wang2020all} detect the oriented rectangular box and transform the oriented rectangular box into a boundary. The boundary is served as the fiducial points for the Thin-Plate-Spline transformation to rectify the irregular text into a regular one. Qiao \etal~\cite{qiao2020text} use a similar way which develops a segmentation detector to generate the fiducial points. ABCNet~\cite{liu2020abcnet} uses a one-stage detector~\cite{tian2019fcos}, which incorporates the parameterized Bezier curves to represent the text instance as well as a new RoI operation (BeizerAlign) for sampling the arbitrarily-shaped text features into the horizontal format. Its improved version~\cite{liu2021abcnetv2} adopts the BiFPN~\cite{tan2020efficientdet} as the backbone and uses an attention-based recognizer to further improve the performance. SwinTextSpotter~\cite{huang2022swintextspotter} further exploits a new way of synergy between the detection and recognition, termed the Recognition Conversion module, to make detection differentiable with recognition loss. TESTR~\cite{zhang2022text} designs a single encoder and dual decoder structure based on the Deformable-DETR~\cite{zhu2020deformable} to remove the hand-designed components. ABINet++~\cite{fang2022abinet++} uses the framework in~\cite{liu2021abcnetv2} and further use a recognizer with an autonomous, bidirectional, and iterative language model~\cite{fang2021read} to improve the performance. 

    \subsection{Point-based and Transcription-only Scene Text Spotter}
    Several recent studies have explored the use of transcription-only data to develop or assist text spotting systems. For instance, TTS~\cite{kittenplon2022towards} incorporates an RNN recognition head in Deformable-DETR and employs the Hungarian algorithm~\cite{kuhn1955hungarian} to enhance the model using solely text transcription annotations. Similarly, TOSS~\cite{tang2022you} utilizes transcription-only data, annotated by voice, to train the model. Benefited from its proposed coarse-to-fine cross-attention mechanism, TOSS can generate coarse text masks without the necessity of detection data.

    Our prior work, the single-point text spotter (SPTS) \cite{peng2022spts} used an auto-regressive Transformer to create long sequences for all text instances, which can also be used for single-point and transcription-only training. However, it faced limitations with inference speed, particularly for images with many text instances. To overcome these challenges, we developed SPTS v2, a model that significantly improves both inference speed and performance by subtly enhancing auto-regressive prediction with enhanced efficiency. 
    
    \begin{figure*}[t!]
    	\centering
    	\includegraphics[width=\linewidth]{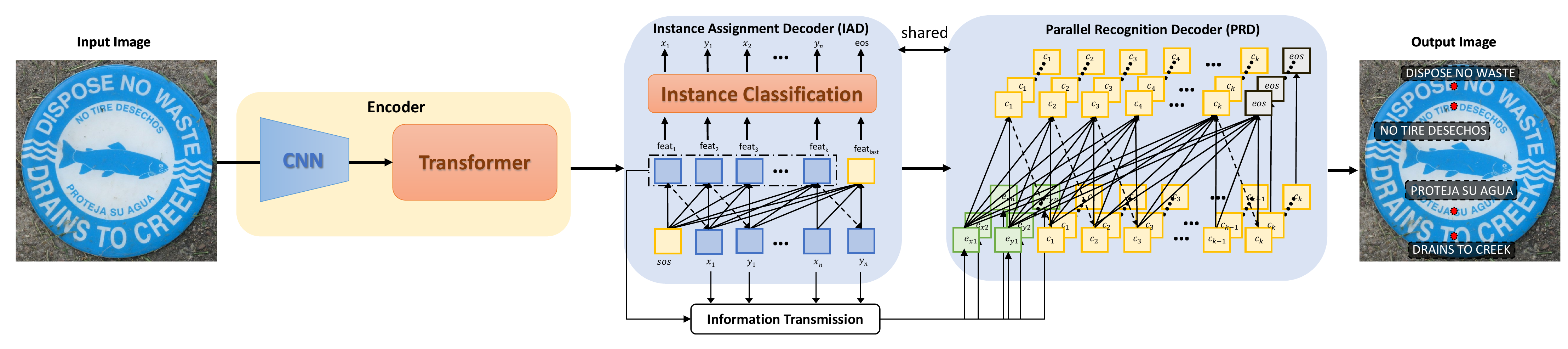}
    	\caption{Overall framework of the proposed SPTS v2. The visual and contextual features are first extracted by a series of CNN and Transformer encoders. Then, the features are auto-regressively decoded into a sequence that contains localization and recognition information through IAD and PRD, respectively. For IAD, it predicts coordinates of all center points of text instances inside the same sequence, while for the PRD, the recognition results are predicted in parallel. Note that IAD shares identical parameters with PRD, and thus no additional parameters are introduced for the PRD stage.}
    	\label{fig:method}
    \end{figure*}

    \section{Methodology}
    \label{sec:method}
    \subsection{Preliminary}
    Most of the existing text spotting algorithms require customized modules to bridge the detection and recognition blocks, where backbone features are cropped and shared between detection and recognition heads, \emph{e.g.}, BezierAlign~\cite{liu2020abcnet}, RoISlide~\cite{feng2019textdragon}, and RoIMasking~\cite{liao2020masktext}. 
    
    Inspired by the Pix2Seq~\cite{chen2021pix2seq}, our prior work, the single-point text spotter (SPTS) \cite{peng2022spts} casts the text spotting problem as a language modeling task, based on an intuitive assumption that if a deep model knows what and where the target is, it can be taught to tell the results by the desired sequence. 
    SPTS used an auto-regressive Transformer for generating long sequences for all text instances, managing unordered instances efficiently. Despite its effectiveness, its inference speed was significantly impacted, especially with images containing numerous text instances. 

    \subsection{SPTS v2}
    To enhance both inference speed and overall performance, the design of SPTS v2 features an Instance Assignment Decoder (IAD) and a Parallel Recognition Decoder (PRD), each tackling distinct aspects of the process. The IAD predicts `indicators' (such as $x$ and $y$ coordinates) auto-regressively for each text instance in the integrated sequence, while the PRD, leveraging these indicators, enables parallel prediction for corresponding text recognition results. Both IAD and PRD leverage a Transformer decoder structure. To achieve parameter reduction, the decoders of both models share parameters and are supervised by the gradients from the detection and recognition tasks. Initially, the shared decoder operates as the IAD, predicting 'indicators'. Subsequently, with the incorporation of a novel information transmission approach, the shared decoder transforms into the PRD, allowing for parallel prediction of text recognition results. This information transmission method merges location embeddings with IAD features, thereby facilitating a gradient flow from recognition features.
    
\begin{figure}[t!]
        \centering
        \begin{minipage}{0.31\linewidth}
            \includegraphics[width=2.6cm, height=1.6cm]{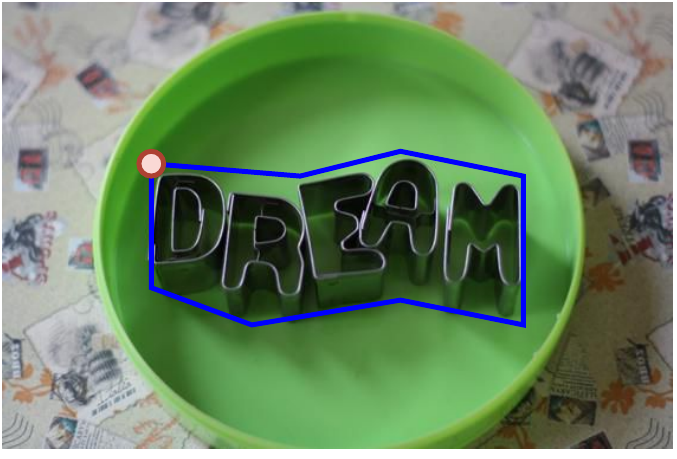}
            \centerline{\small{(a) Top-left}}
            \label{fig:tl}
        \end{minipage}
        \begin{minipage}{0.31\linewidth}
            \includegraphics[width=2.6cm, height=1.6cm]{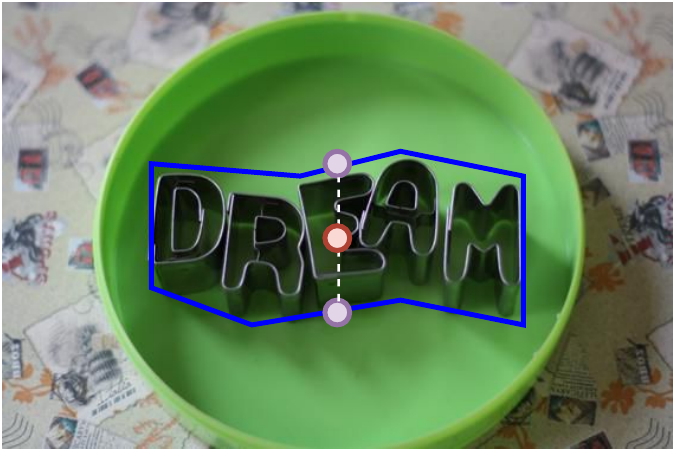}
            \centerline{\small{(b) Central}}
            \label{fig:ctr}
        \end{minipage}
        \begin{minipage}{0.31\linewidth}
            \includegraphics[width=2.6cm, height=1.6cm]{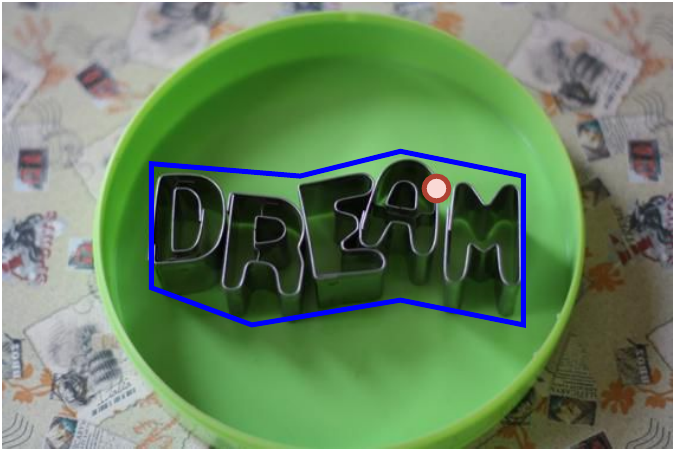}
            \centerline{\small{(c) Random}}
            \label{fig:rand}
        \end{minipage}
        \caption{Indicated points using different positions.}
        \label{fig:pts_pos}
    \end{figure}

    Specifically, as shown in Fig.~\ref{fig:method}, each input image is first encoded by CNN and Transformer encoders to extract visual and contextual features. Then, the captured features are decoded by a Transformer decoder, where tokens are predicted in an auto-regressive manner. Unlike previous algorithms, we further simplify the bounding box to a corner point located at the upper left of the first character or the center of the text instance, described in Fig.~\ref{fig:pts_pos}, in the text instance. Benefiting from such a simple yet effective representation, the modules carefully designed based on prior knowledge, such as grouping strategies utilized in segmentation-based methods and feature sampling blocks equipped in box-based text spotters, can be eschewed.

    \subsection{Sequence Construction}
    To express the target text instances by a sequence, it is required to convert the continuous descriptions (\emph{e.g.}, bounding boxes) to a discretized space. To this end, we simplify the bounding box to a single point and use the variable-length transcription instead of the single-token object category. 
    
    The main limitation of SPTS~\cite{peng2022spts} is that the long sequence length will significantly slow down the inference speed. This is because the recognition results normally are fixed to the maximum length equal to 25 and 100 for word-level and line-level text instances, respectively. To this end, in SPTS v2, we design the Instance Assignment Decoder (IAD) and Parallel Recognition Decoder (PRD) to overcome such limitations.

    \begin{figure}[t!]
        \centering
        \includegraphics[width=0.8\linewidth]{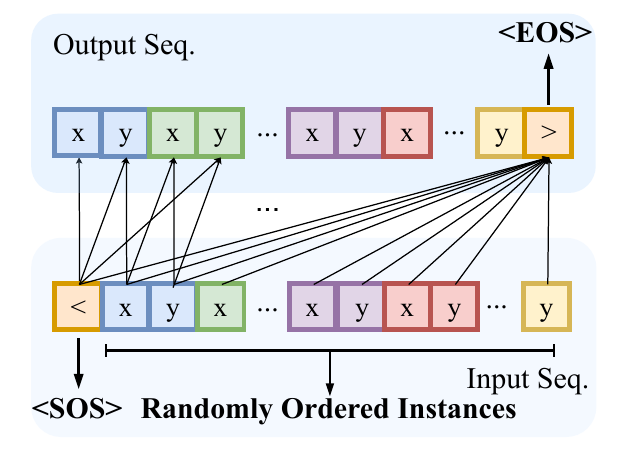}
        \caption{Input and output sequences of the Instance Assignment Decoder (IAD). Every group of x and y represents different text instance.}
        \label{fig:input_output_seq}
    \end{figure}
    
    \begin{figure}[t!]
        \centering
        \includegraphics[width=1.0\columnwidth]{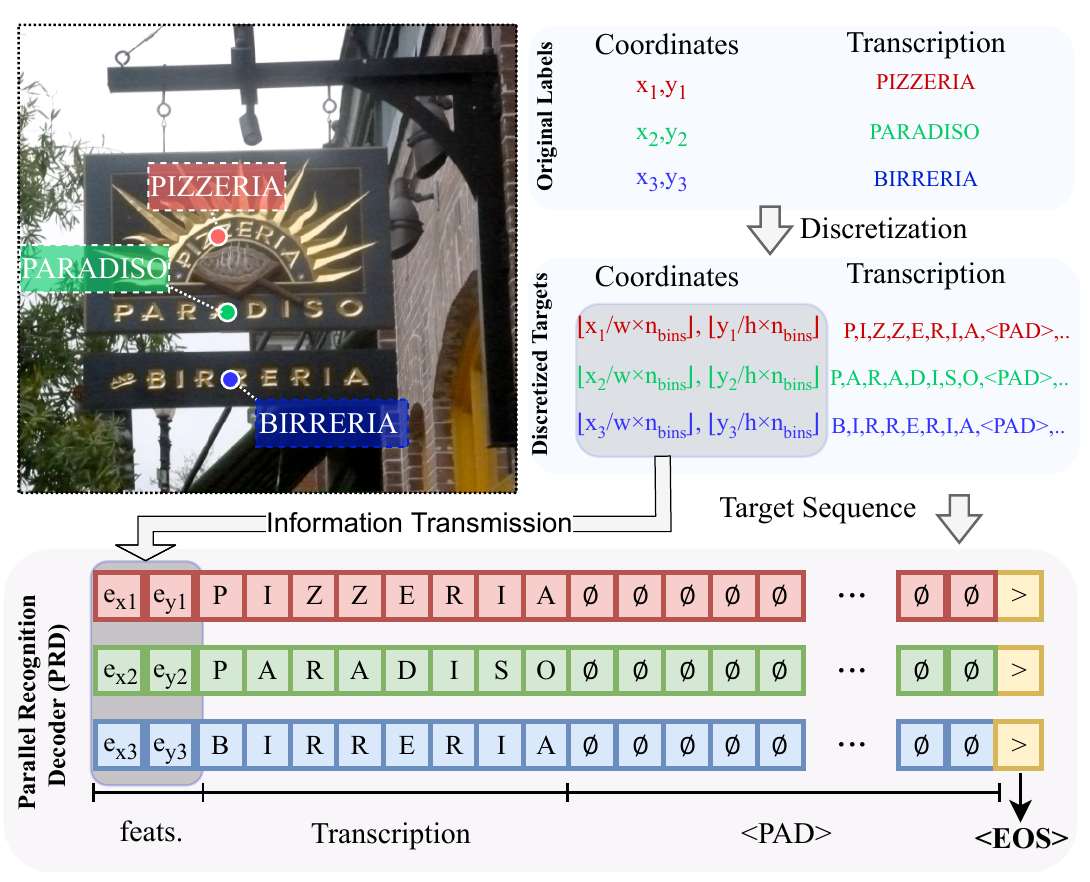}
        \caption{Input and output sequences of the Parallel Recognition Decoder (PRD). Each line represents different text instance. Given the features generated from information transmission, the recognition results are predicted in parallel until reaching the maximum length or the EOS symbols.}
        \label{fig:prd}
    \end{figure}
    
    \subsection{Instance Assignment Decoder} 
    \label{subsec:IAD}
    The auto-regressive decoder is known to be effective in literature~\cite{chen2021pix2seq,peng2022spts}; however, it is intuitive that this is a time-consuming solution given the long sequence of the text instances. To improve efficiency, the SPTS v2 divides detection and recognition into a two-stage workflow by sharing the same Transformer decoder. The first stage is called Instance Assignment Decoder (IAD). In the first stage, SPTS v2 only decodes the center point for every text instance until the end of the sequence comes. An intuitive pipeline is shown in Fig.~\ref{fig:input_output_seq}. 
    
    Specifically, the continuous coordinates of the central point of the text instance are uniformly discretized into integers between $[1, n_{bins}]$, where $n_{bins}$ controls the degree of discretization. For example, an image with a long side of 800 pixels requires only $n_{bins}=800$ to achieve zero quantization error. Note that the central point of the text instance is obtained by averaging the upper and lower midpoints as shown in Fig.~\ref{fig:pts_pos} (b). As so far, a text instance can thereby be represented by a sequence of three parts, \emph{i.e.}, $[x, y, t]$, where $(x, y)$ are the discretized coordinates and $t$ is the transcription text that will be predicted in PRD. Notably, the transcriptions are inherently discrete, \emph{i.e.}, each of the characters represents a category. 
    
    $<$SOS$>$ and $<$EOS$>$ tokens are inserted into the head and tail of the sequence, indicating the start and the end of a sequence, respectively. Therefore, given an image that contains $N$ text instances, the constructed sequence will include $2N$ discrete tokens, where the text instances would be randomly ordered. In fact, as shown in previous works~\cite{chen2021pix2seq}, the randomly ordered text instances can be effectively learned, and thus it achieves the label assignment for different hidden features inconspicuously, which subtly avoids an explicit label assignment like using bipartite matching that plays a vital role for the DETR series~\cite{carion2020end,zhu2020deformable,meng2021-CondDETR,liu2022dabdetr}. In fact, compared with other label assignments, the instance assignment is intuitively more efficient. The dense label assignment methods~\cite{liao2019mask,liu2020abcnet} use the non-maximum suppression (NMS) to select the suitable detection results for recognition. The bipartite matching label assignment methods~\cite{zhang2022text,kittenplon2022towards} use a maximum number of instances to detect and recognize texts, which consumes additional computation for the empty text instances.

    \subsection{Parallel Recognition Decoder}
    With the help of the IAD, we separate the different text instances. The content of different text instances will be obtained at the same time in the Parallel Recognition Decoder. Different from the generic object detection that categorizes objects into fixed categories, recognizing the text content is a sequence classification problem that has a variable length of the target sequence. This may cause misalignment issues and can consume more computational resources. To eliminate such problems, we first pad or truncate the texts to a fixed length $K$, where the $<$PAD$>$ token is used to fill the vacancy for shorter text instances. In addition, supposing there are $n_{cls}$ categories of characters (\emph{e.g.}, 97 for English characters and symbols), the vocabulary size of the dictionary used to tokenize the sequence can be calculated as $n_{cls} + 3$, where the extra three classes are for $<$PAD$>$, $<$SOS$>$, and $<$EOS$>$ tokens. Empirically, we set the $K$ and $n_{bins}$ to 25 (or 100 for SCUT-CTW1500) and 1,000, respectively, in our experiments. Moreover, the maximum value of $n_{ti}$ is set to 60, which means the sequence containing more than 60 text instances will be truncated. An illustration of the PRD is shown in Fig.~\ref{fig:prd}.

    We assume that one image includes $N$ text instances, and every instance includes the maximum number of $K$ characters. It takes $N_{v1}$ loops for SPTS to predict this image, where $N_{v1}$ is defined as:
    \begin{equation}
        N_{v1} = (2+K) \cdot N+1.
    \end{equation}
    While for SPTS v2, it only needs $N_{v2}$ for-loops, where $N_{v2}$ is:
    %
    \begin{equation}
        N_{v2} = 2\cdot N+K+1,
    \end{equation}
    with a $K\cdot (N-1)$ reduction. 
    In our implementation, $N$ and $K$ are set to 60 and 25, respectively. In this case, SPTS requires 1,621 auto-regressive loops while SPTS v2 requires only 146 loops, with 91.0\% (1475/1621) reduction rate of the number of loops. Actually, inside the PRD, SPTS v2 can also take an early end if all instances have met the end of sequence symbol. Through PRD, the inference speed can be significantly improved.

    \subsection{Information Transmission} 
    The parameters of the above two decoders are shared and supervised by the detection and recognition gradients. 
    However, there is information loss between different text instances. 
    In the conference version~\cite{peng2022spts}, the information of the previously detected text can be sensed by the recognition token and the gradient of text recognition can be passed on to supervise the predictions of different text instances. Such interaction is also important for the Parallel Recognition Decoder in SPTS v2 to find the correct position of the text. To address this issue, we propose an information transmission method. Formally, we first extract the hidden text instance location features (short for $feat.$ in Fig.~\ref{fig:method}) and the corresponding prediction results of the text location (\emph{e.g.}, $x_1, y_1$). Then, we convert the text instance location results into an embeddings which is then added to the text instance location features. The process can be formulated as follow: 
    \begin{equation}
        embed_{x_i} = embedding(x_i),\  i=0,1,2,...n.
    \end{equation}
    \begin{equation}
        embed_{y_i} = embedding(y_i), \  i=0,1,2,...n.
    \end{equation}
    \begin{equation}
        e_{x_i} = feat_{x_i} + embed_{x_i},
        \label{eq:3}
    \end{equation}
    \begin{equation}
        e_{y_i} = feat_{y_i} + embed_{y_i}.
    \end{equation}
    With the help of the information transmission, the gradient of later text recognition can be passed on to different text instances by the $feat_{x_i}$ or $feat_{y_i}$, and the information of the previously detected text can be sensed by the recognition token in PRD stage through the features. PRD takes these prior information as the first two queries to instruct the decoder and thus recognize all the text instances in parallel, as shown in Fig.~\ref{fig:prd}. Such a straightforward transmission is essential to the SPTS v2 in practice.

    \subsection{Model Training}
    \def\bw{{\bf w}}
    \def\bs{{\bf s}}
    \def\bI{{\bf I}}
    Since the SPTS v2 is trained to predict tokens, it only requires to maximize the likelihood loss at training time, which can be written as:
    \begin{equation}
    \label{eq_objective}
    \mathcal{L} = {\rm max} \sum_{i=1}^{L} {\bw}_i \log
     P(\tilde{{\bs}}_i | {\bI}, {\bs}_{1:i}),
    \end{equation}
    where $\textbf{I}$ is the input image, $\tilde{\textbf{s}}$ is the output sequence, $\textbf{s}$ is the input sequence, $ L$ is the length of the sequence, and $\textbf{w}_i$ is the weight of the likelihood of the $i$-th token, which is empirically set to 1. For both IAD and PRD, they share the same Transformer and require only the cross-entropy loss, maintaining a concise pipeline. 

    \subsection{Inference}
    
    At the inference stage, SPTS v2 first sequentially predicts the tokens of location in IAD until the end of the sequence token $<$EOS$>$ occurs. Then, the information transmission will integrate the detection features to auto-regressively predict the text contents in parallel. The predicted sequence will subsequently be divided into multiple segments. Therefore, the tokens can be easily translated into point coordinates and transcriptions, yielding the text spotting results. In addition, the likelihood of all tokens in the corresponding segment is averaged and assigned as a confidence score to filter the original outputs, effectively removing redundant and false-positive predictions.

    \section{Experiments}\label{sec:exp}
    
    We report the experimental results on five benchmarks, including horizontal dataset ICDAR 2013~\cite{karatzas2013icdar}, multi-oriented dataset ICDAR 2015~\cite{karatzas2015icdar}, arbitrarily shaped datasets Total-Text~\cite{ch2017total} and SCUT-CTW1500~\cite{liu2019curved}, and Inverse-Text~\cite{ye2022dptext} dataset. 
    
    \subsection{Datasets} 
        
    {\bf Curved Synthetic Dataset 150k.} 
    It is admitted that the performance of text spotters can be improved by pre-training on synthesized samples. Following previous work~\cite{liu2020abcnet}, we use the 150k synthetic images generated by the SynthText~\cite{gupta2016synthetic} toolbox, which contains around one-third of curved texts and two-thirds of horizontal instances.
    
    {\bf ICDAR 2013}~\cite{karatzas2013icdar} 
    contains 229 training and 233 testing samples, while the images are primarily captured in a controlled environment, where most of the texts are horizontally presented and explicitly focused.

    {\bf ICDAR 2015}~\cite{karatzas2015icdar} 
    consists of 1,000 training and 500 testing images that were incidentally captured, containing multi-oriented text instances presented in complicated backgrounds with strong variations in blurring, distortions, etc.

    {\bf Total-Text}~\cite{ch2017total} 
    includes 1,255 training and 300 testing images, where at least one curved sample is presented in each image and annotated with polygonal bounding boxes at the word level.
        
    {\bf SCUT-CTW1500}~\cite{liu2019curved} 
    is another widely used benchmark designed for spotting arbitrarily shaped scene text, which involves 1,000 and 500 images for training and testing. The text instances are labeled by polygons at the text-line level.
    
    {\bf Inverse-Text}~\cite{ye2022dptext} 
    is a recently proposed dataset focused on arbitrary-shape scene text with about 40\% inverse-like instances, containing 500 testing images. Following the previous work~\cite{ye2022dptext}, we test this dataset with the model trained on Total-Text.
    
    \subsection{Evaluation Protocol}
    \label{subsec:eval_protocol}

\begin{figure}[t!]
        \centering
        \includegraphics[width=0.8\columnwidth]{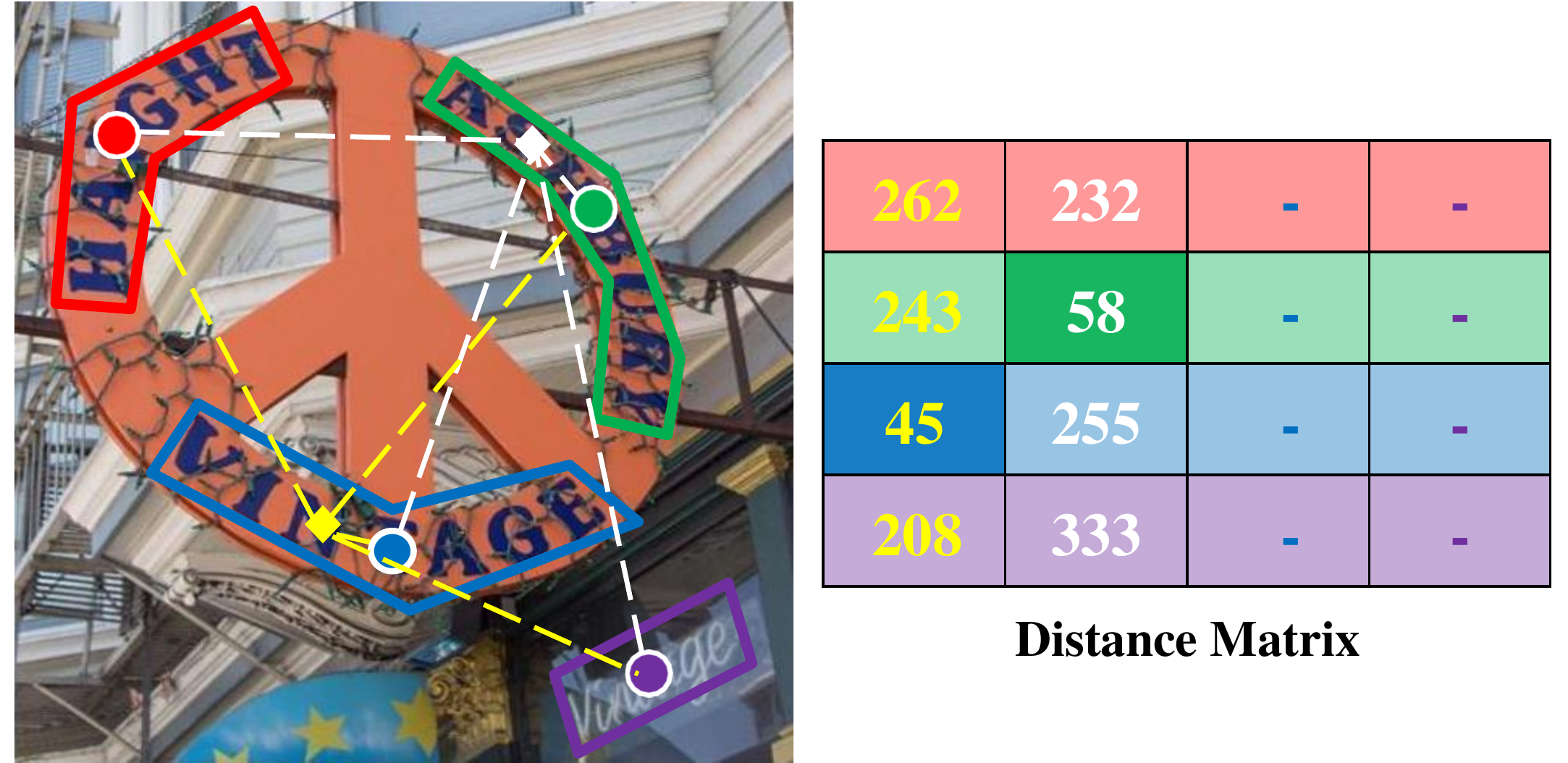}
        \caption{Illustration of the point-based evaluation metric. Diamonds are predicted points, and circles represent ground truth.}
        \label{fig:eval}
    \end{figure}

    The existing evaluation protocol of text spotting tasks consists of two steps. Firstly, the intersection over union (IoU) scores between ground-truth (GT) and detected boxes are calculated; and only if the IoU score is larger than a designated threshold (usually set to 0.5), the boxes are matched. Then, the recognized content inside each matched bounding box is compared with the GT transcription; only if the predicted text is the same as the GT will it contribute to the end-to-end accuracy. However, in the proposed method, each text instance is represented by a single point; thus, the evaluation metric based on the IoU is not available to measure the performance. Meanwhile, comparing the localization performance between bounding-box-based methods and the proposed point-based methods might be unfair, \emph{e.g.}, directly treating points inside a bounding box as true positives may overestimate the detection performance. To this end, we propose a new evaluation metric to ensure a relatively fair comparison to existing approaches, which mainly considers the end-to-end accuracy as it reflects both detection and recognition performance (failure detections usually lead to incorrect recognition results). Specifically, as shown in Fig.~\ref{fig:eval}, we modified the text instance matching rule by replacing the IoU metric with a distance metric, \emph{i.e.}, the predicted point that has the nearest distance to the central point of the GT box would be selected, and the recognition results will be measured by the same full-matching rules used in existing benchmarks. Only one predicted point with the highest confidence will be matched to the ground truth; others are then marked as false positives. 
        
    To explore whether the proposed evaluation protocol can genuinely represent the model accuracy, Tab.~\ref{tab:ab_metric} compares the end-to-end recognition accuracy of ABCNetv1~\cite{liu2020abcnet} and ABCNetv2~\cite{liu2021abcnetv2} on Total-Text~\cite{ch2017total} and SCUT-CTW1500~\cite{liu2019curved} under two metrics, \emph{i.e.}, the commonly used bounding box metric that is based on IoU, and the proposed point-based metric. The results demonstrate that the point-based evaluation protocol can well reflect the performance, where the difference between the values evaluated by box-based and point-based metrics is no more than 0.5\%. For example, the ABCNetv1 model achieves 53.5\% and 53.0\% scores on the SCUT-CTW1500 dataset under the two metrics, respectively. Therefore, we use the point-based metric to evaluate the proposed SPTS v2 in the following experiments.
    
    \begin{table}[t!]
        \centering
        \caption{Comparison of the end-to-end recognition performance evaluated by the proposed point-based metric and polygon-based metric. Results are reproduced using official codes.}
        \label{tab:ab_metric}
        \small
        \begin{tabular}{r|cc|cc}\hline
            \multirow{2}{*}{Method} & \multicolumn{2}{c|}{Total-Text} & \multicolumn{2}{c}{SCUT-CTW1500} \\ \cline{2-5}
             & \multicolumn{1}{c|}{Box} & Point & \multicolumn{1}{c|}{Box} & \multicolumn{1}{c}{Point} \\ \hline
            ABCNet v1~\cite{liu2020abcnet} & 67.2 & 67.4 & 53.5 & 53.0  \\
            ABCNet v2~\cite{liu2021abcnetv2} & 71.7 & 71.9 & 57.6 & 57.1 \\
            \hline
        \end{tabular}
    \end{table}

    \subsection{Implemented Details}
    The model is first pretrained on a combination dataset that includes Curved Synthetic Dataset 150k~\cite{liu2020abcnet}, MLT-2017~\cite{nayef2017icdar2017}, ICDAR 2013~\cite{karatzas2013icdar}, ICDAR 2015~\cite{karatzas2015icdar}, and Total-Text~\cite{ch2017total} for 150 epochs, which is optimized by the AdamW~\cite{loshchilov2017decoupled} with an initial learning rate of $5\times 10^{-4}$, while the learning rate is linearly decayed to $1\times 10^{-5}$. After pre-training, the model is then fine-tuned on the training split of each target dataset for another 200 epochs, with a fixed learning rate of $1\times 10^{-5}$. The entire model is distributively trained on 16 NVIDIA A100 GPUs with a batch size of 8 per GPU. Note that the effective batch size is 64 because two independent augmentations are performed on each image in a mini-batch, following \cite{chen2021pix2seq,hoffer2020augment}. In addition, we use ResNet-50 as the backbone network, while both the Transformer encoder and decoder consist of 6 layers with eight heads. Regarding the architecture of the Transformer, we adopt the Pre-LN Transformer~\cite{xiong2020layer}.
    During training, the short size of the input image is randomly resized to a range from 640 to 896 (intervals of 32) while keeping the longer side shorter than 1,600 pixels, following to previous methods. Random cropping and rotating are employed for data augmentation. At the inference stage, we resize the short edge to 1,000 while keeping the longer side shorter than 1,824 pixels, following the previous works~\cite{zhang2022text,liu2020abcnet,liu2021abcnetv2}.
    
    To enhance the accuracy and reliability of our model's output, we have instituted a three-step procedure to filter out redundant predictions. First, we derive classification confidence scores for both detection and recognition tokens. Then, we compute the mean confidence score across these tokens. In the final step, we set a threshold based on this average score, filtering out predictions that fall below it. Through this approach, most of the false positives can be reduced.

    \subsection{Ablation Study} 
    \label{subsec:abs}
    
    \subsubsection{Ablation Study of Designs}
    We first conduct ablation studies to evaluate different designs of SPTS v2. Because the PRD requires started tokens to predict the recognition results in parallel, at least the hidden features (termed Feat) or the embeddings of the locations (termed Token) are required in the baseline setting. The results are shown in Tab.~\ref{tab:abla_cores}. We can see that without sharing the parameters of IAD and PRD, the performance encounters a 1.4\% reduction in terms of the Full metric of the Total-Text dataset. In addition, according to lines 1, 2, and 4 of the table, integrating the Token and Feat can further improve the performance, \emph{e.g.}, 3\% and 5.5\% higher than independently using the Token and Feat, respectively, in terms of the Full metric. The results demonstrate the importance of the information transmission. We use the pre-trained model to test the results. 

    \begin{table}[t!]
    \centering
        \caption{Ablation studies on Total-Text \emph{w.r.t.}\  the designs.  ``None" represents lexicon-free. ``Full" represents that we use all the words that appeared in the test set. The Feat and token represent the items on the right of Equation~\ref{eq:3}, respectively. Shared represents sharing the parameters of the IAD and PRD.}
        \small
        \label{tab:abla_cores}
        \begin{tabular}{r|c|c|c|cc}
        
        \hline
        \multirow{2}{*}{Method}       & \multirow{2}{*}{Token} & \multirow{2}{*}{Feat}  & \multirow{2}{*}{Shared} & \multicolumn{2}{c}{Total-Text } \\ \cline{5-6} 
                                       &                      &      &                          & \multicolumn{1}{c|}{None}      & Full     \\ \hline
        Baseline             &    \checkmark                  &    & \checkmark                           & \multicolumn{1}{c|}{66.4}      & 78.4     \\  
        Baseline             &                      & \checkmark    &  \checkmark                        & \multicolumn{1}{c|}{66.1}      & 75.9     \\ 
        Baseline             &  \checkmark      & \checkmark  &                             & \multicolumn{1}{c|}{68.1}      & 80.0  \\
        Baseline             &  \checkmark            & \checkmark  &  \checkmark                           & \multicolumn{1}{c|}{\textbf{68.5}}      & \textbf{81.4}     \\ \hline
    \end{tabular}
    \end{table}
    
    \subsubsection{Ablation Study of the Position of The Indicated Point}
    Intuitively, all points in the region enclosed by the bounding box should be able to represent the target text instance. To explore the differences, we conduct ablation studies that use three different strategies to get the indicated points (see Fig.~\ref{fig:pts_pos}), \emph{i.e.}, the \emph{central} point obtained by averaging the upper and lower midpoints, the \emph{top-left} corner, and the \emph{random} point inside the box. It should be noted that we use the corresponding ground-truth here to calculate the distance matrix for evaluating the performance to ensure the fair comparison, \emph{i.e.}, the distance to the ground-truth top-left point is used for \emph{top-left}, the distance to the ground-truth central point for \emph{central}, and the closest distance to the ground-truth polygon for \emph{random}. 
    
    \begin{table}[t!]
        \centering
        \caption{Ablation study of the position of the indicated point.}
        \label{tab:ab1_position_point}
        \small
        \begin{tabular}{c|c|c|c|cc}\hline
        \multirow{2}{*}{Position} & \multicolumn{2}{c|}{E2E Total-Text} & \multicolumn{2}{c}{E2E SCUT-CTW1500} \\ \cline{2-5} 
         & \multicolumn{1}{c|}{None} & Full & \multicolumn{1}{c|}{None} & \multicolumn{1}{c}{Full} \\ \hline
        Central & \textbf{74.2} & \textbf{82.4} & \textbf{63.6} & \textbf{83.8} \\
        Top-left & 71.6 & 79.7 & 61.4 & 82.0 \\
        Random & 73.2 & 80.8 & 62.3 & 81.1 \\ \hline
        \end{tabular}
    \end{table}

    The results are shown in Tab.~\ref{tab:ab1_position_point}, where the result of \emph{left-top} is the worst. The result of \emph{random} is close to \emph{central} with approximately 1\% worse in terms of the None metric. Although the central point shows the best performance against other formats, it suggests that the performance is not very sensitive to the positions of the point annotation.

    \begin{table}[t!]
    \centering
    \caption{Comparison with different shapes of bounding box. $N_{p}$ is the number of parameters required to describe the location of text instances by different representations. 2x twices the training schedule. }
    \label{tab:ab_box_shape}
    \small
    \begin{tabular}{r|cc|cc|c}
    \hline
    \multirow{2}{*}{Variants} & \multicolumn{2}{c|}{Total-Text}                   & \multicolumn{2}{c|}{SCUT-CTW1500} &
    \multirow{2}{*}{$N_{p}$} \\ \cline{2-5} 
                            & \multicolumn{1}{c|}{None}    & \multicolumn{1}{c|}{Full} 
                            & \multicolumn{1}{c|}{None}    & \multicolumn{1}{c|}{Full}
                            & \\ \hline
    SPTS-Bezier & 60.6 & 71.6 & 52.6 & 73.9 & 16 \\ 
    SPTS-Bezier 2x & 62.9 & 74.4 & 51.1 & 74.3 & 16 \\ 
    SPTS-Rect & 71.6 & 80.4 & 62.2 & 82.0 & 4 \\  
    SPTS-Non-Point & 64.7 & 71.9 & 55.4 & 74.3 & 0 \\  
    SPTS-Point &\textbf{74.2} & \textbf{82.4} & \textbf{63.6} & \textbf{83.8} & 2  \\ \hline
    SPTS v2-Bezier & 63.2 & 73.6 & 52.3 & 70.2 & 16 \\ 
    SPTS v2-Rect & 72.6 & 79.5 & 55.0 & 71.5 & 4 \\
    SPTS v2-Point &\textbf{75.5} & \textbf{84.0} & \textbf{63.6} & \textbf{84.3} & 2  \\ \hline
    \end{tabular}
    \end{table}

    \subsubsection{Comparison Between Different Representations}
    The proposed method can be easily extended to produce bounding boxes by modifying the point coordinates to bounding box locations during sequence construction. Here, we conduct ablation studies to explore the influence by only changing representations of the text instances. 
    Specifically, four variants are explored, including 1) the Bezier curve bounding box; 2) the rectangular bounding box; 3) the indicated point; and 4) non-point.
    Note for the non-point representation, we only implement the results using SPTS, because it is hard to implement using SPTS v2, which requires the prediction of the location for the PRD stage. 
    
\begin{figure}[t!]
        \centering
        \begin{minipage}[c]{0.85\linewidth}
            \centering
            \includegraphics[width=7.2cm, height=2.0cm]{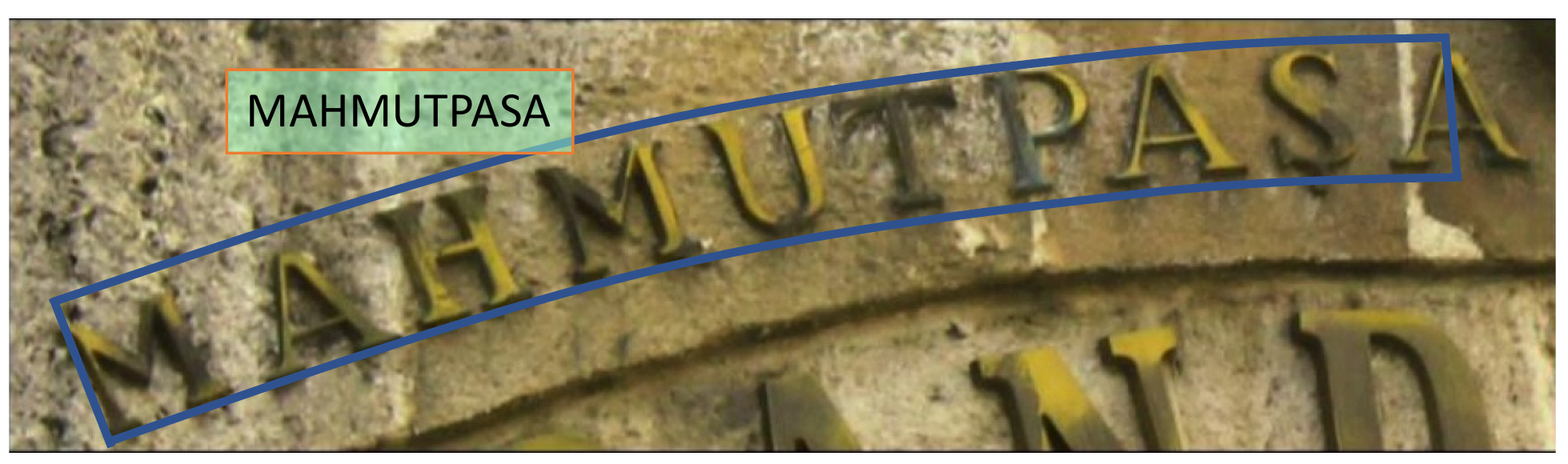}
        \end{minipage}
        
        \begin{minipage}[c]{0.85\linewidth}
            \centering
            \includegraphics[width=7.2cm, height=2.0cm]{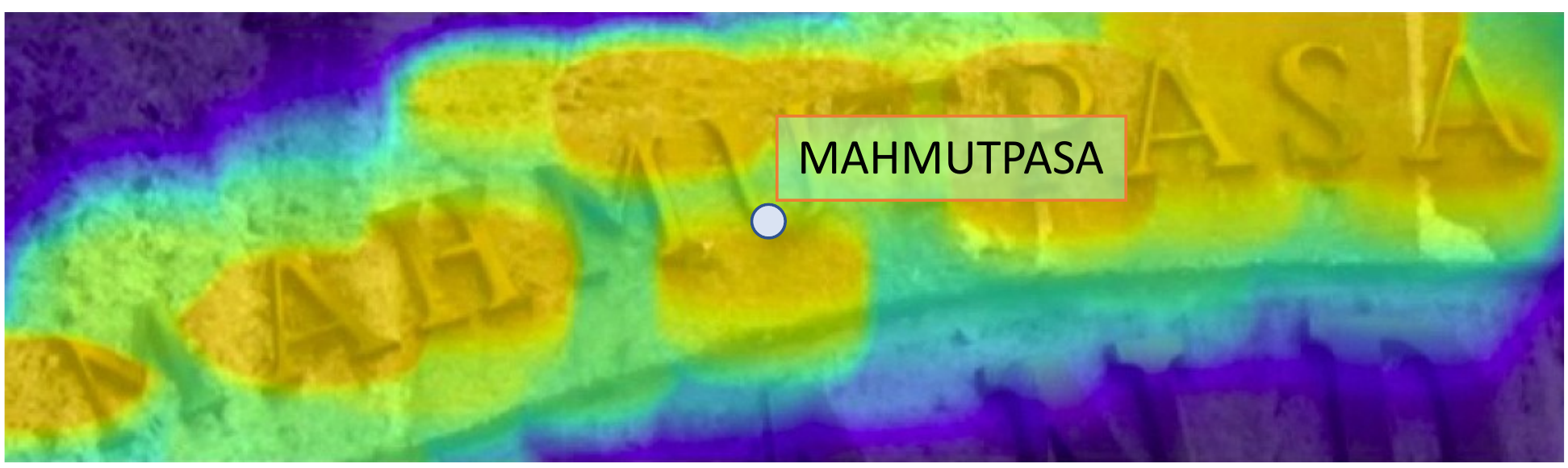}
        \end{minipage}
        \caption{The receptive field can be beneficial for the final recognition. Upper: the result of ABCNet v2. Lower: rough receptive field of our method. }
        \label{fig:recept}
    \end{figure}

    Since we only focus on end-to-end performance here, to minimize the impact of the detection results, each method uses corresponding representations to match the GT box in the evaluation. That is, the single-point model uses the evaluation metrics introduced in Sec.~\ref{subsec:eval_protocol}, \emph{i.e.}, distance between points; the predictions of SPTS v2-Rect are matched to the circumscribed rectangle of the polygonal annotations; the SPTS v2-Bezier adopts the original metric that matches polygon boxes; and the evaluation metric for non-point can be referred to Sec.~\ref{subsec:non-point}.
    As shown in Tab.~\ref{tab:ab_box_shape}, the SPTS v2-point achieves the best performance on both the Total-Text and SCUT-CTW1500 datasets, outperforming the other representations by a large margin. Such experimental results suggest that a low-cost annotation, \emph{i.e.}, the indicated point, is capable of providing supervision for the text spotting task. 
    Here, to safely ground such findings, we further provide analysis as follows:
    \begin{itemize}
    \item The results of SPTS-Rect and SPTS-Bezier are obtained using the same training schedule as SPTS-Point. To further explore if the former may require a longer training schedule, we compare the SPTS-Bezier trained for $2\times$ epochs with SPTS-Point in Tab.~\ref{tab:ab_box_shape}. It can be seen that the SPTS-Bezier with $2\times$ epochs does not significantly outperform the counterpart with $1\times$ epochs and is still inferior to the SPTS-Point with $1\times$ epochs. In addition, using a longer schedule even results in lower performance on SCUT-CTW1500 for SPTS-Bezier in terms of the None metric, which suggests the training schedule may not be the case. 
    \item To further eliminate the influence of the different metrics, we also directly adopt the center point inside the rectangular or Bezier-curved bounding box to test the same point metric as our method. The results are shown in Tab.~\ref{tab:point_in_box}, which show that the variance is still consistent with the conclusion of Tab.~\ref{tab:ab_metric}, \emph{i.e.}, the result of the point metric is close to that of the box or polygon-based metrics in terms of the None metric. 
    \item As we can observe from previous scene text spotting method~\cite{liu2021abcnetv2}, sometimes the recognition results can still be accurate even if the detection result is inaccurate, like missing some of the regions of the characters, as shown in the top of Fig.~\ref{fig:recept}. This is because the alignment for text recognition is based on the feature space, in which the cropped features have enough receptive fields for the text contents. Such phenomenon can also support our finding: as shown in the bottom of Fig.~\ref{fig:recept}, because the image is globally encoded in our method, an approximate location could be enough for the model to capture the desired features in vicinity, which may further release the power of the Transformer. 
    \end{itemize}

    \subsubsection{Order of Text Instances}
    As described in Sec.~\ref{sec:method}, the text instances are randomly ordered in the constructed sequence. Here, we further investigate the impact of the order of text instances. The performances on Total-Text and SCUT-CTW1500 of different ordering strategies are presented in Tab.~\ref{tab_ordering}. The ``Area'' and ``Dist2ori'' mean that text instances are sorted by the area and the distance to the top-left origin in descending order, respectively. The ``Top-down'' indicates that text instances are arranged from top to bottom. It can be seen that the random order for our method achieves the best performance.
    Although the result may be counter-intuitive that the randomly ordered setting achieves the best; however, it is consistent to the result of pix2seq model, which struggles with missing objects when using a set order. A random order in such case could potentially resolve this as it might catch those missed objects later. This empirically makes the model more robust due to the different sequences constructed for the same image at different iterations.
    
    \begin{table}[t]
        \centering
        \caption{Ablation study of different ordering strategies of text instances in the sequence construction.}
        \label{tab_ordering}
        \small
        \begin{tabular}{r|cc|cc}
            \hline
            \multirow{2}{*}{Order} & \multicolumn{2}{c|}{Total-Text} & \multicolumn{2}{c}{SCUT-CTW1500} \\
            \cline{2-5}
             & None & Full & None & Full \\
            \hline
            Area & 70.7 & 79.2 & 59.0 & 75.3 \\
            Topdown & 73.2 & 81.3 & 62.7 & 79.7 \\
            Dist2ori & 72.1 & 81.8 & 61.1 & 79.6 \\
            \hline 
            Random & \textbf{74.2} & \textbf{82.4} & \textbf{63.6} & \textbf{83.8}  \\
            \hline
        \end{tabular}
    \end{table}
    
    \begin{table}[t]
        \centering
        \caption{Comparison of the end-to-end recognition performance evaluated by the proposed point-based metric and box-based metric.}
        \label{tab:point_in_box}
        \small
        \begin{tabular}{r|cc|cc}
            \hline
            \multirow{2}{*}{Order} & \multicolumn{2}{c|}{Total-Text} & \multicolumn{2}{c}{SCUT-CTW1500} \\
            \cline{2-5}
             & None & Full & None & Full \\
            \hline
            boxes & 72.6 & 79.5 & 55.0 & 71.5 \\
            boxes-point & \textbf{72.9} & \textbf{81.1} & \textbf{56.5} & \textbf{77.7} \\ \hline
            polygon & 63.2 & 73.6 & 52.3 & 70.2 \\
            polygon-point & \textbf{64.9} & \textbf{76.0} & \textbf{52.6} & \textbf{79.5} \\
            \hline
        \end{tabular}
    \end{table}

    \begin{table}[t!]
    \centering
    \caption{End-to-end recognition results and detection results on Total-Text. ``None"
    represents lexicon-free. ``Full" represents that we use all the words that appeared in the test set. Decoder 1 represents using one layer for the decoder instead of using six layers.}
    \label{tab:num_decoder}
    \small
    \begin{tabular}{r|c|c}
    \hline
    \multirow{2}*{Method} & \multicolumn{2}{c}{Total-Text End-to-End} \\ \cline{2-3}
    & None                      & \multicolumn{1}{c}{Full} \\ \hline   
    R18 decoder 1 & 11.5 & 26.0 \\
    R18 & 60.6 & 74.3    \\ 
    R34 decoder 1 & 22.7 & 49.0 \\
    R34 & 64.7 & 76.7 \\
    R50 decoder 1 & 53.7 & 65.5 \\
    R50 & 68.5 & 81.4 \\
    \hline
    \end{tabular}
    \end{table}
    
    \begin{table}[t!]
    \centering
    \caption{Ablation study on simulated noisy annotated data results on SCUT-CTW1500.  ``None" represents lexicon-free. ``Full" represents that we use all the words appeared in the test set.}
    \label{tab:noise_e2e}
    \small
    \begin{tabular}{r|c|c}
    \hline
    \multirow{2}*{Method} & \multicolumn{2}{c}{SCUT-CTW1500 End-to-End}  \\ \cline{2-3} 
               & None                      & Full  \\ \hline
    \multicolumn{3}{c}{Original} \\ \hline 
    ABCNet v2 \cite{liu2021abcnetv2}  & 57.5       & 77.2   \\ 
    SPTS v2      &  63.6       & 84.3 \\ \hline 
    \multicolumn{3}{c}{Disturbance radius of 5} \\ \hline 
    ABCNet v2 \cite{liu2021abcnetv2}  & 55.1      & 76.0    \\
    SPTS v2 & 63.0 & 82.1 \\  \hline
    \multicolumn{3}{c}{Disturbance radius of 10} \\ \hline 
    ABCNet v2 \cite{liu2021abcnetv2}  & 54.9      & 75.4    \\
    SPTS v2 & 61.6 & 81.1 \\ 
    \hline
    \end{tabular}
    \end{table}
    
    \begin{figure}[ht!]
        \centering
        \begin{minipage}[c]{0.47\linewidth}
            \centering
            \includegraphics[width=4.3cm, height=2.0cm]{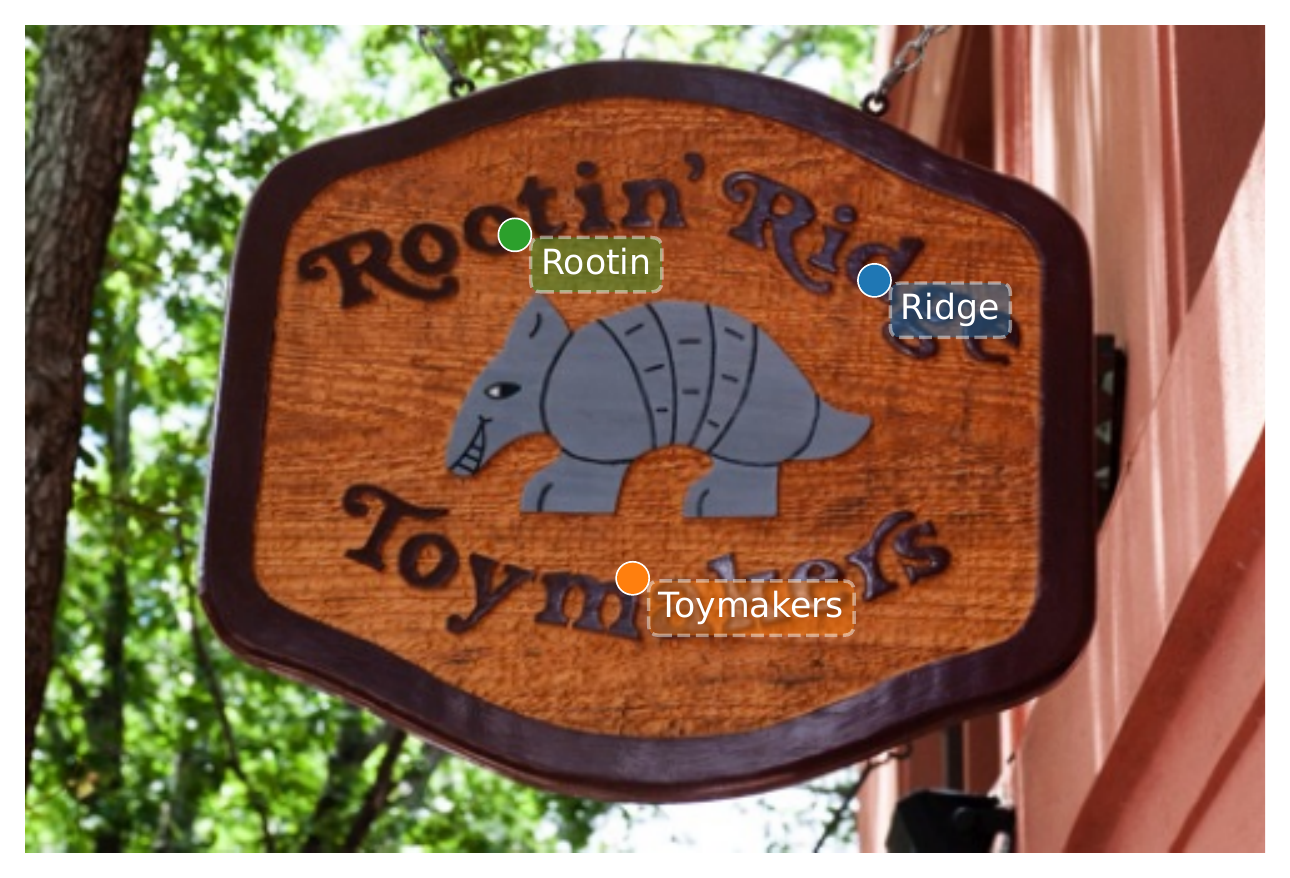}
        \end{minipage}
        \begin{minipage}[c]{0.47\linewidth}
            \centering
            \includegraphics[width=4.3cm, height=2.0cm]{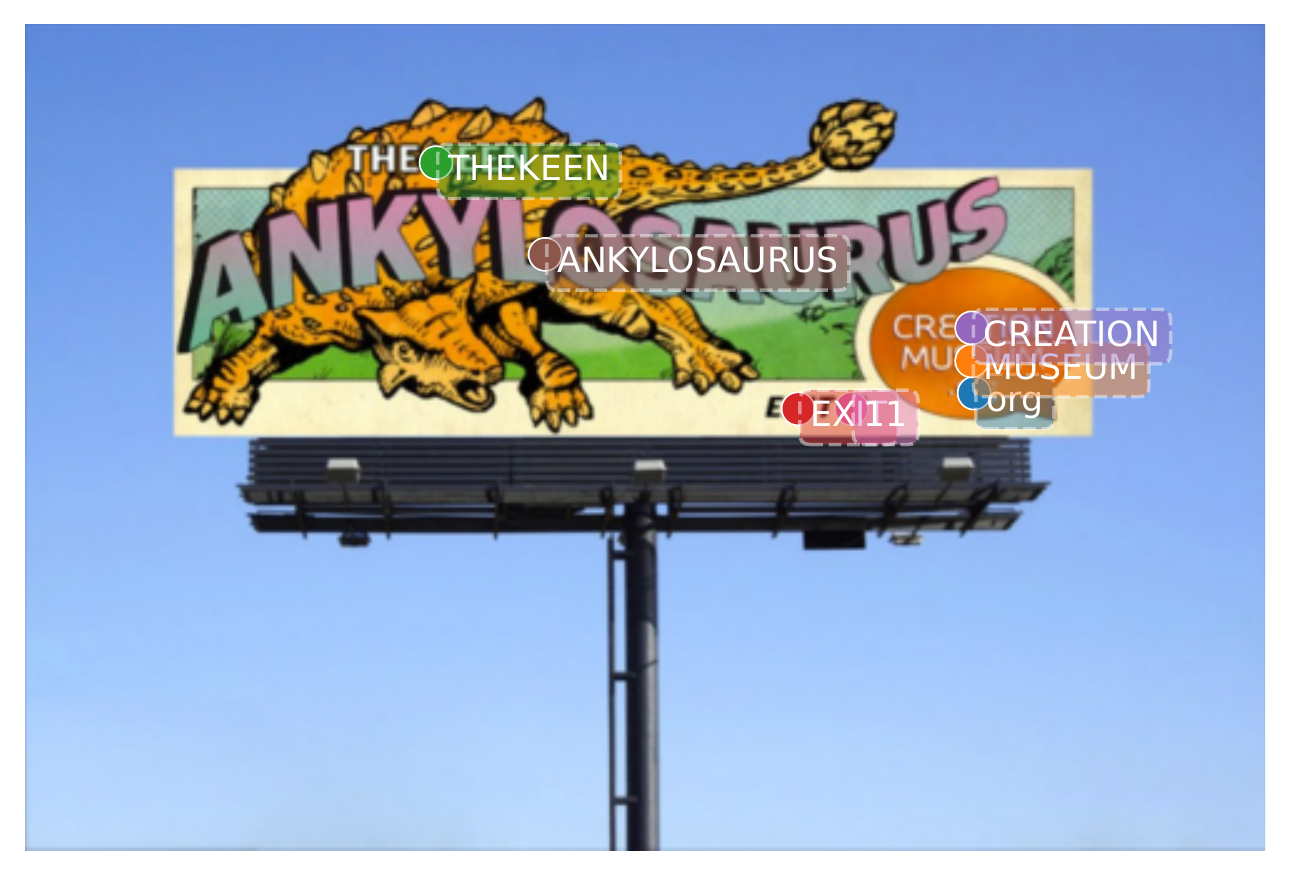}
        \end{minipage}
        \begin{minipage}[c]{0.47\linewidth}
            \centering
            \includegraphics[width=4.3cm, height=2.0cm]{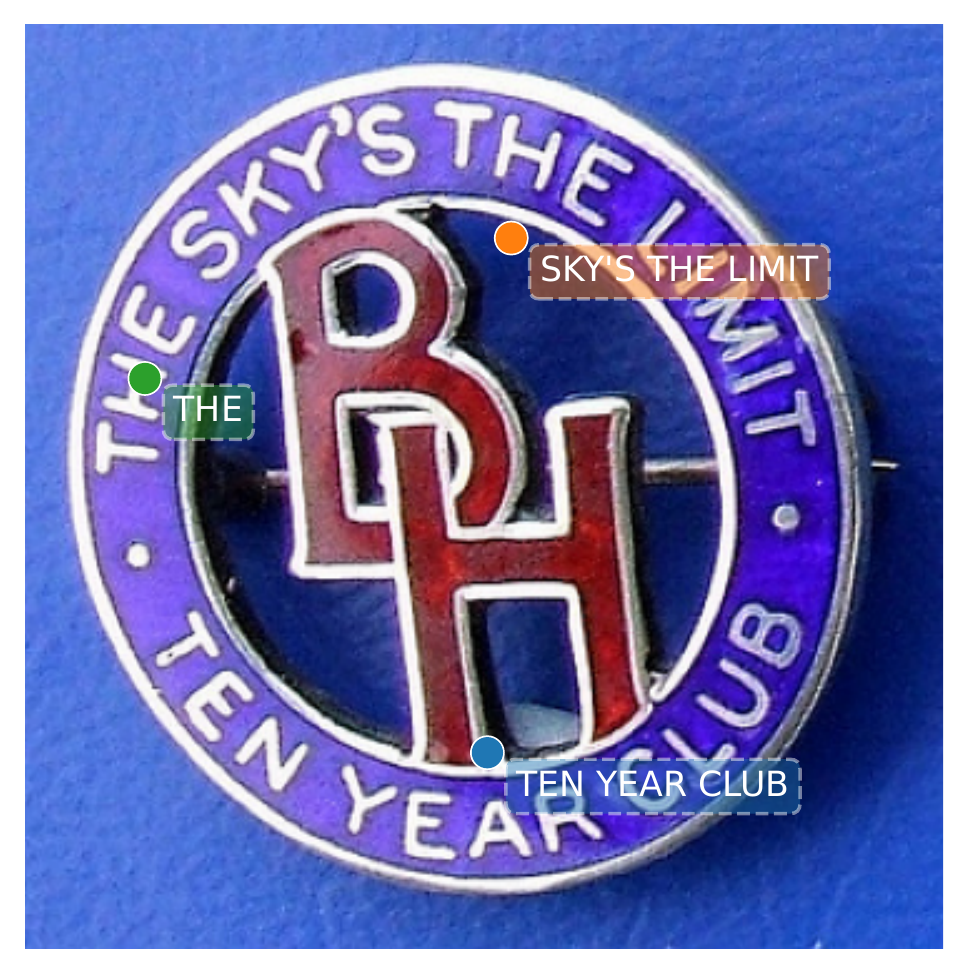}
        \end{minipage}
        \begin{minipage}[c]{0.47\linewidth}
            \centering
            \includegraphics[width=4.3cm, height=2.0cm]{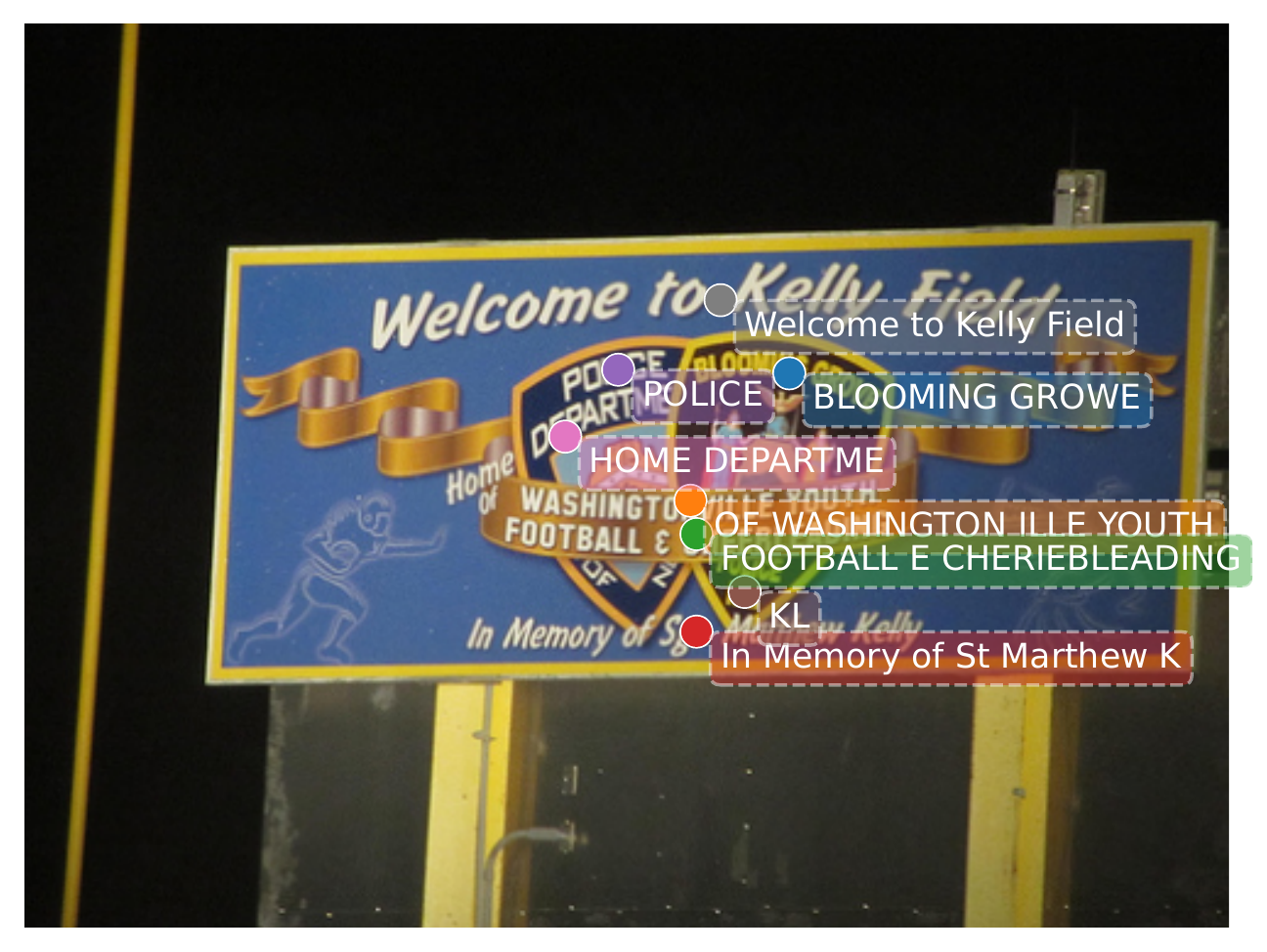}
        \end{minipage}
        \begin{minipage}[c]{0.47\linewidth}
            \centering
            \includegraphics[width=4.3cm, height=2.0cm]{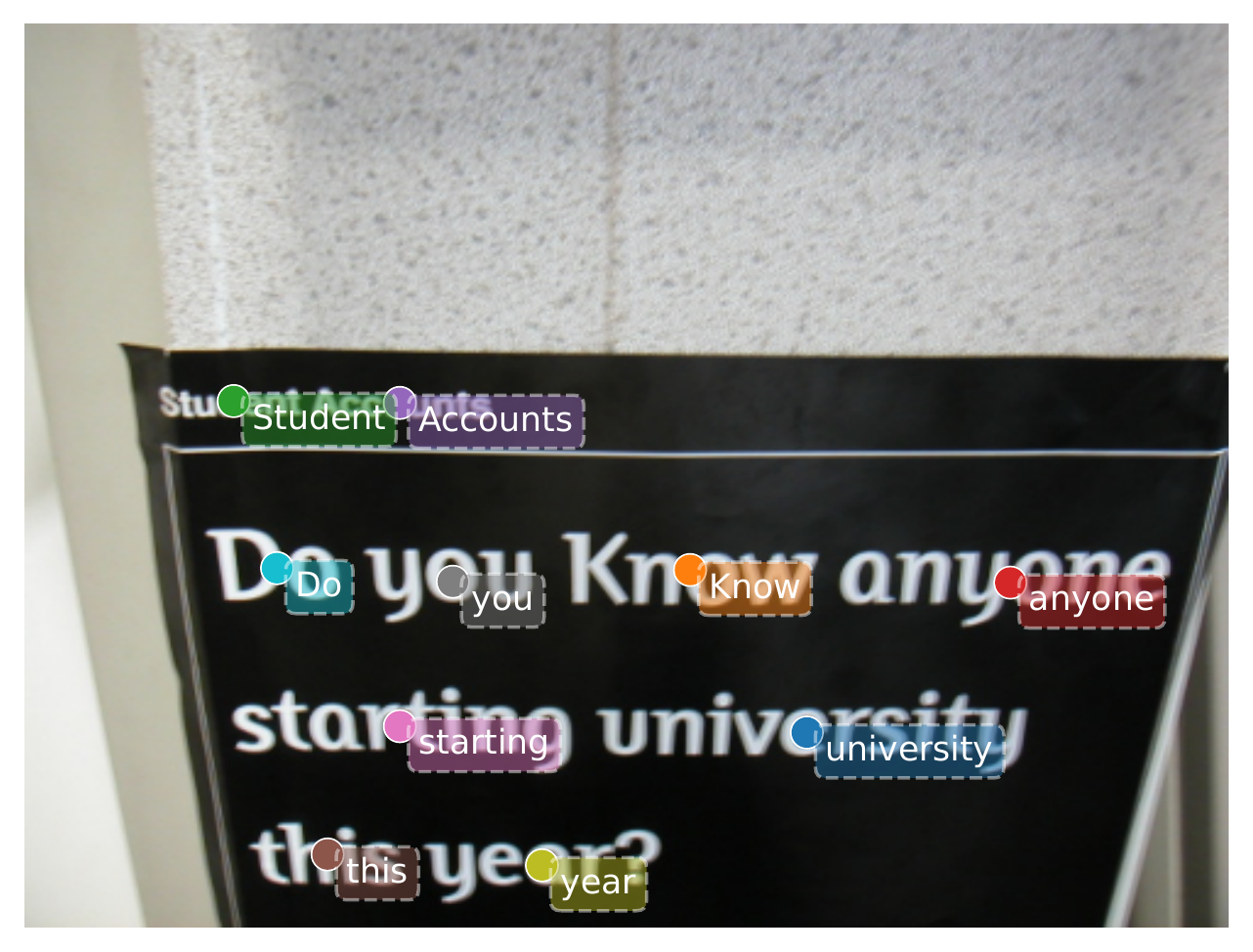}
        \end{minipage}
        \begin{minipage}[c]{0.47\linewidth}
            \centering
            \includegraphics[width=4.3cm, height=2.0cm]{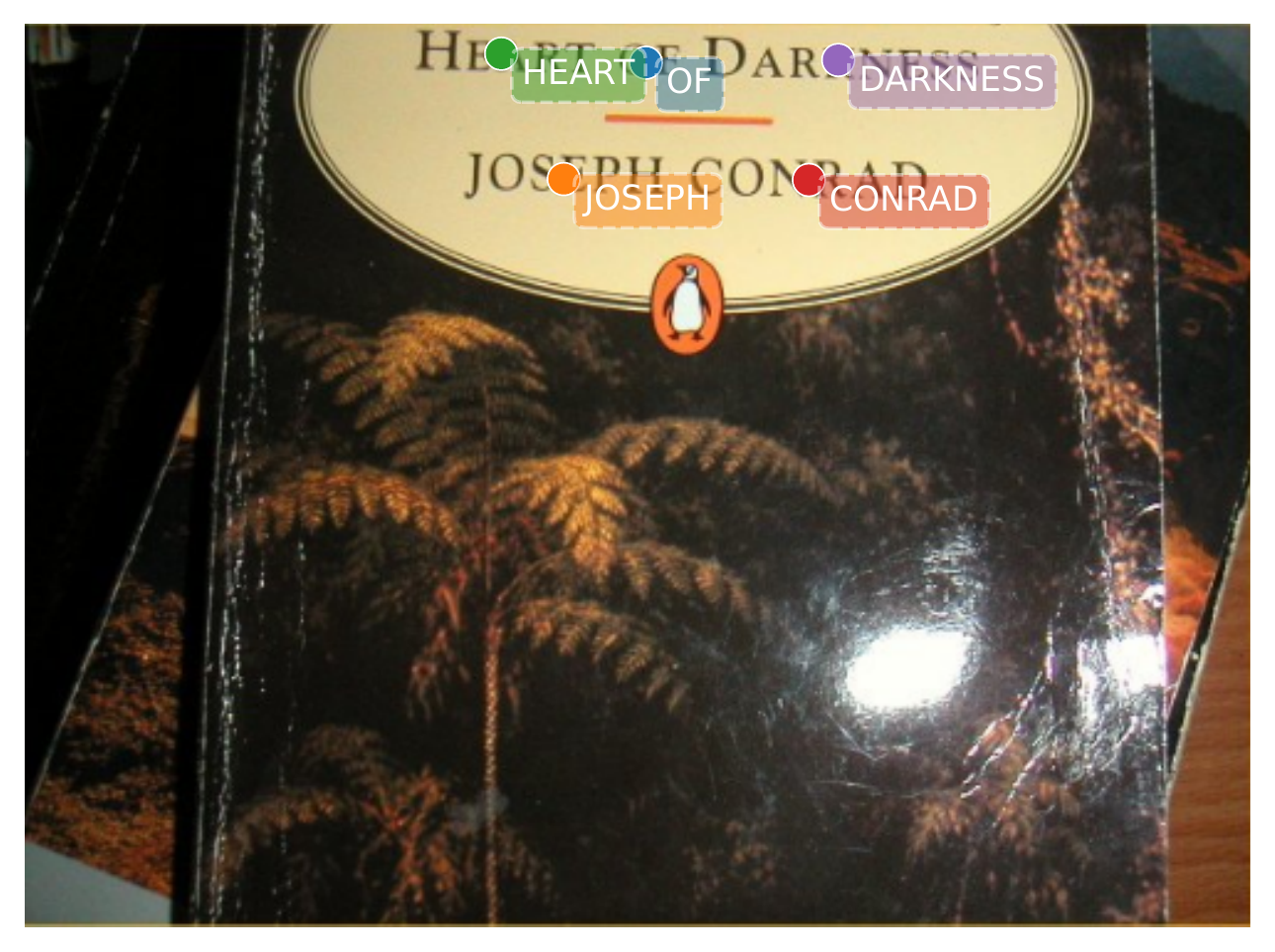}
        \end{minipage}
        \begin{minipage}[c]{0.47\linewidth}
            \centering
            \includegraphics[width=4.3cm, height=2.0cm]{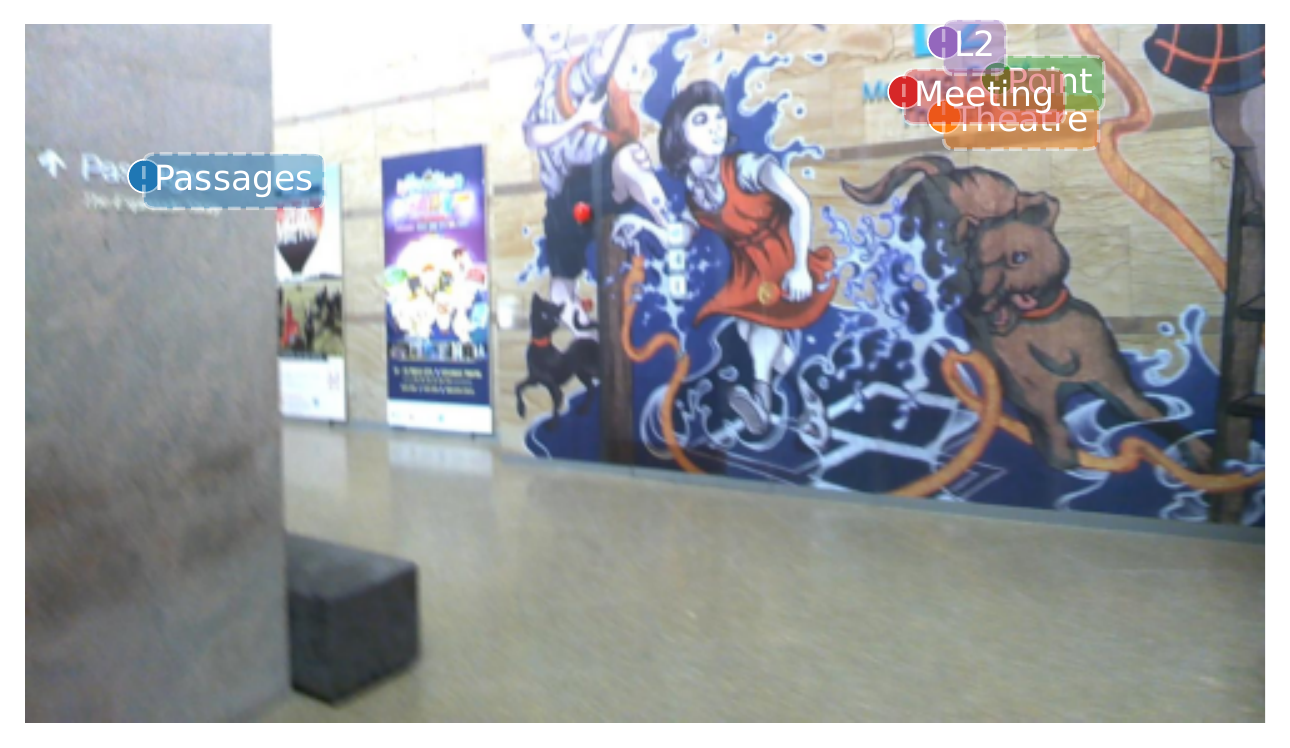}
        \end{minipage}
        \begin{minipage}[c]{0.47\linewidth}
            \centering
            \includegraphics[width=4.3cm, height=2.0cm]{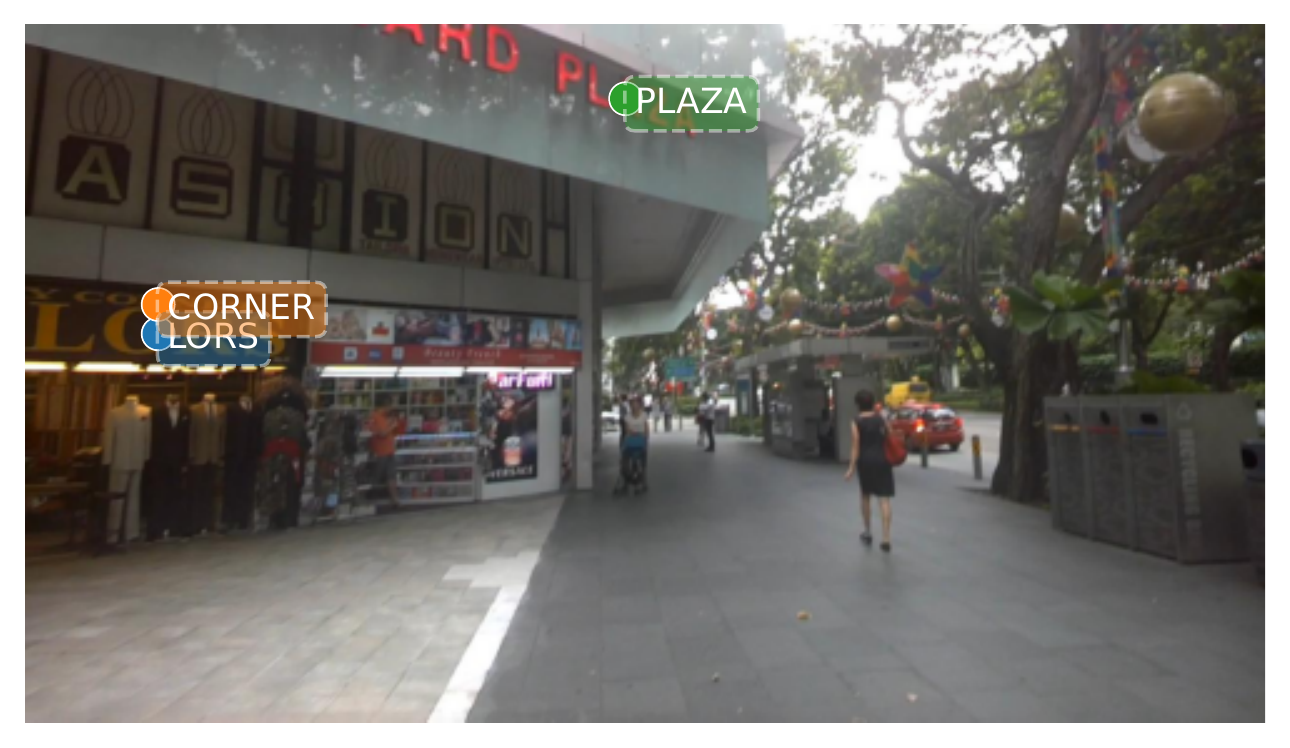}
        \end{minipage}
        \begin{minipage}[c]{0.47\linewidth}
            \centering
            \includegraphics[width=4.3cm, height=2.0cm]{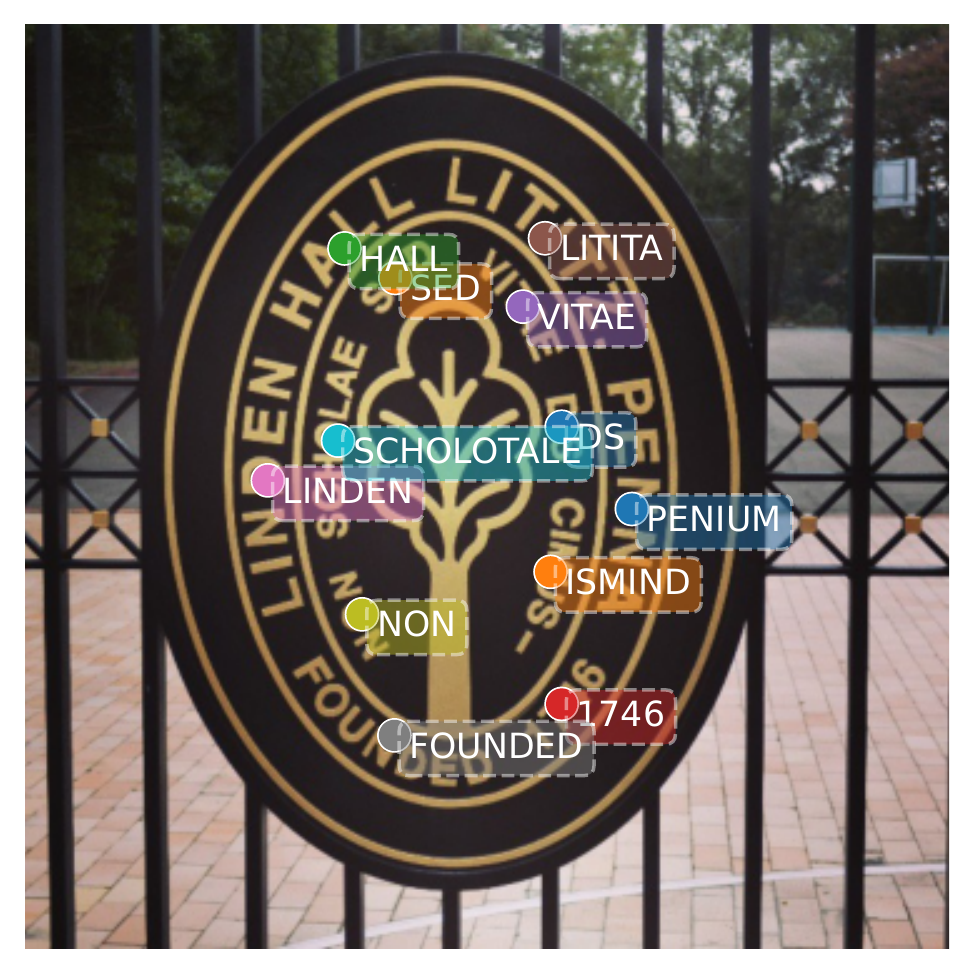}
        \end{minipage}
        \begin{minipage}[c]{0.47\linewidth}
            \centering
            \includegraphics[width=4.3cm, height=2.0cm]{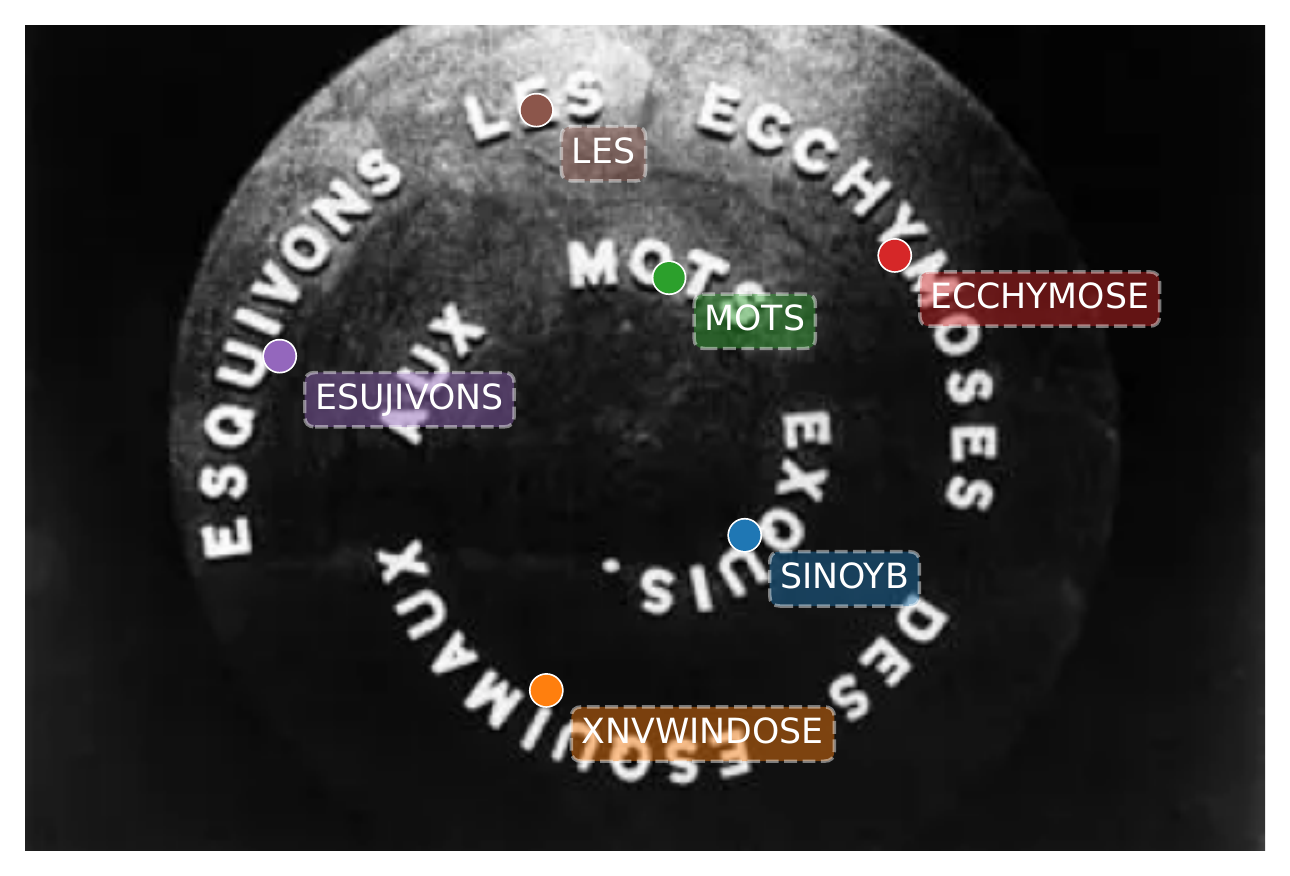}
        \end{minipage}
        \caption{Qualitative results on the scene text benchmarks. Images are selected from Total-Text (first row), SCUT-CTW1500 (second row), ICDAR 2013 (third row), ICDAR 2015 (fourth row), and Inverse-Text (fifth row). 
        Best viewed on screen.
        }\label{fig:qual}
    \end{figure}

    \begin{table*}[!t]\small
      \centering
      \caption{End-to-end text spotting results on Total-Text, SCUT-CTW1500, ICDAR2015, and Inverse-Text. `None' means lexicon-free. `Full' indicates that we use all the words appeared in the test set. `S', `W', and `G' represent recognition with `Strong', `Weak', and `Generic' lexicon, respectively.}
      \newcommand{\tabincell}[2]{\begin{tabular}{@{}#1@{}}#2\end{tabular}}
      \begin{tabular}{@{} r | cc | cc | ccc | cc @{}}
        \hline
        \multirow{2}*{Methods}  & \multicolumn{2}{c|}{Total-Text} & \multicolumn{2}{c|}{SCUT-CTW1500} &
        \multicolumn{3}{c|}{ICDAR 2015 End-to-End}  &  \multicolumn{2}{c}{Inverse-Text} \\
        \cline{2-10}
            & None & Full & None & Full  & S & W & G & None & Full  \\
        \hline
        \multicolumn{10}{c}{Bounding Box-based methods} \\ \hline
        Mask TextSpotter \cite{liao2019mask}  &65.3&77.4  &--&-- &83.0&77.7&73.5 &39.0 &43.5   \\
        Unconstrained   \cite{qin2019towards}  &67.8&-- &--&--  &--&-- &-- &--&-- \\
        CharNet \cite{xing2019convolutional}  & 66.2 &--  &--&-- &80.1&74.5&62.2 &--&-- \\
        FOTS \cite{liu2018fots} &--&-- &21.1&39.7  &83.6&79.1&65.3 &--&-- \\
        TextDragon \cite{feng2019textdragon} &48.8&74.8 &39.7&72.4  &82.5&78.3&65.2 &--&-- \\
        Text Perceptron \cite{feng2019textdragon} &69.7 & 78.3 &57.0 & -- &80.5 &76.6 &65.1 &--&-- \\
        ABCNet \cite{liu2020abcnet} &64.2&75.7 &45.2&74.1  &--&-- &-- &22.2 &34.3\\
        Boundary TextSpotter \cite{wang2020all} &65.0 & 76.1 &--&-- &79.7&75.2&64.1 &--&-- \\
        Mask TextSpotter v3 \cite{liao2020masktext} &71.2&78.4 &--&--  &83.3&78.1&74.2 &--&-- \\
        PGNet \cite{wang2021pgnet}  & 63.1 & --  & -- & --  & 83.3 & 78.3 & 63.5  &--&-- \\
        MANGO \cite{qiao2021mango}  & 72.9 & 83.6  & \textbf{58.9} & 78.7  & 81.8 & 78.9 & 67.3 &--&-- \\
        ABCNet v2 \cite{liu2021abcnetv2}  & 70.4 & 78.1  & 57.5 & 77.2  & 82.7 & 78.5 & 73.0 & 34.5 & 47.4 \\
        PAN++ \cite{liu2021abcnetv2}  & 68.6 & 78.6  & -- & --  & 82.7 & 78.2 & 69.2 &--&-- \\
        TESTR \cite{zhang2022text}  & 73.3 & 83.9  & 56.0 & \textbf{81.5}  & 85.2 & 79.4 & 73.6 & 34.2 & 41.6 \\
        SwinTextSpotter \cite{huang2022swintextspotter}  & 74.3 & 84.1  & 51.8 & 77.0  & 83.9 & 77.3 & 70.5 & \textbf{55.4} & \textbf{67.9} \\
        TTS \cite{kittenplon2022towards}  & 78.2 & \textbf{86.3}  & -- & --  & 85.2 & 81.7 & \textbf{77.4} &--&-- \\
        GLASS \cite{ronen2022glass}  & \textbf{79.9} & 86.2  & -- & --  & 84.7 & 80.1 & 76.3 &--&-- \\
        Boundary TextSpotter'22 \cite{luboundary}  & 66.2 & 78.4  & 46.1 & 73.0  & 82.5 & 77.4 & 71.7 &--&-- \\
        SRSTS \cite{wu2022decoupling}  & 78.8 & 86.3  & -- & --  & \textbf{85.6} & \textbf{81.7} & 74.5 &  - & - \\
        \hline
        \multicolumn{10}{c}{Point-based methods} \\ \hline
        TOSS~\cite{tang2022you} & 65.1 & 74.8 & 54.2 & 65.3
        & 65.9 & 59.6 & 52.4 & - & - \\
        SPTS~\cite{peng2022spts}   & 74.2 & 82.4 & 63.6  & 83.8  & 77.5 & 70.2 & 65.8 &38.3&46.2 \\
        SPTS v2 & \textbf{75.5} & \textbf{84.0} & \textbf{63.6} & \textbf{84.3} & \textbf{82.3} & \textbf{77.7} & \textbf{72.6} & \textbf{63.4} & \textbf{74.9} \\
        \hline
      \end{tabular}
    \label{tab:cp_sota_e2e}
    \end{table*}
    
    \begin{table}[t!]
        \centering
        \caption{End-to-end recognition results on ICDAR 2013. ``S'', ``W'', and ``G'' represent recognition with ``Strong'', ``Weak'', and ``Generic'' lexicon, respectively. SPTS v2 is 19x faster than SPTS.}
        \label{ICDAR 2013 End-to-End recognition result}
        \small
        \begin{tabular}{r|c|c|c|c|c}
        \hline 
        \multirow{2}{*}{Method} & \multicolumn{3}{c|}{IC13 End-to-End}   & \multirow{2}{*}{Para.} & \multirow{2}{*}{FPS} \\ \cline{2-4} 
                                & \multicolumn{1}{c|}{S}    & \multicolumn{1}{c|}{W}    & G  && \\ \hline
        \multicolumn{6}{c}{Bounding Box-based methods} \\ \hline                        
        Jaderberg \textit{et al}.\  \cite{jaderberg2016reading} & 86.4 & -- & -- & -- & -- \\ 
        Textboxes \cite{liao2017textboxes} & 91.6 & 89.7 & 83.9& -- & -- \\ 
        Deep Text Spotter \cite{busta2017deep} & 89.0 & 86.0 & 77.0& -- & -- \\ 
        Li \textit{et al.} \cite{li2017towards} & 91.1 & 89.8 & 84.6 & -- & -- \\ 
        MaskTextSpotter \cite{lyu2018mask} & 92.2 & 91.1 & 86.5 & 45.5M & 4.8 \\ 
        MANGO ~\cite{qiao2021mango} & 93.4 & \textbf{92.3} & \textbf{88.7} & -- & -- \\
        \hline 
        \multicolumn{6}{c}{Point-based methods} \\ \hline
        \methodName\ & 93.3 & 91.7 & 88.5 & 36.5M & 0.4 \\ 
        SPTS v2 & \textbf{93.9} & 91.8 & 88.6 & 36.0M & 7.6 \\
        \hline
        \end{tabular}
    \end{table}
    
    \subsubsection{Ablation Study of Different Settings}
    We further conduct ablation studies \emph{w.r.t.} depth of the sharing decoder layers of both IAD and PRD and various backbones for our framework on Total-Text, as shown in Tab.~\ref{tab:num_decoder}. We observe that using ResNet-34 as backbone surpasses ResNet-18 by 4.1\% in terms of the None metric. Using ResNet-50 as backbone can outperform ResNet-34 by a further 3.8\%. In addition, we find that the number of decoder layers may greatly influence the performance for different backbones. For example, with ResNet-18, ResNet-34, and ResNet-50 as backbones, decreasing the number of decoder layers from 6 to 1 leads to consistent 49.1\%, 42\%, and 15.2\% performance declining in terms of the None metric for the Total-Text dataset.
    
    \subsubsection{Robustness on Noise Data}
    In this section, we conduct experiments on the simulated noisy annotated data to further demonstrate the potential of SPTS v2. For ABCNet v2, we begin by introducing noise to the training data of CTW1500 through random perturbations of the ground truth polygon point annotations. Random perturbations were applied to the ground truth training annotations of polygon points on the CTW1500 dataset. For SPTS v2, we compute the center point of the ground truth polygon point annotations first. Following a similar procedure, we apply random perturbations to the center point on the CTW1500 dataset. Subsequently, both ABCNet v2 and SPTS v2 were trained using the perturbed annotations. The results, as shown in Tab.~\ref{tab:noise_e2e}, reveal that SPTS v2 demonstrates a degradation of approximately 0.6\%, whereas ABCNet v2 exhibits a degradation of approximately 2.4\%, with a disturbance radius of 5, which serves to emphasize the robustness of SPTS v2 to noise. Additionally, with a disturbance radius of 10, SPTS v2 experiences a decline of around 2.0\%, while ABCNet v2 shows a decrease of about 2.6\%. We note that many text instances possess a short height, and a disturbance radius of 10 for the center point might cause it to fall outside of the text instances.

    \subsection{Comparison with Existing Methods on Scene Text Benchmarks}

    \footnotetext{\sf  https://github.com/aim-uofa/AdelaiDet}
    
    \subsubsection{Horizontal-Text Dataset}
    Tab.~\ref{ICDAR 2013 End-to-End recognition result} compares the proposed method with existing methods on the widely used ICDAR 2013~\cite{karatzas2013icdar} benchmark. Our method achieves the best performance under the ``strong'' lexicon while achieving comparable performance on ``weak'' and ``generic'' metrics. Note SPTS v2 achieves 20x faster than the previous state-of-the-art single-point-based method with fewer parameters. 
    
    \subsubsection{Multi-Oriented Dataset}\label{sec:multi-orientd dataset}
    The quantitative results of the ICDAR 2015~\cite{karatzas2015icdar} dataset are shown in Tab.~\ref{tab:cp_sota_e2e}. A performance gap between the proposed method and state-of-the-art methods can be found. The proposed method can not accurately recognize tiny texts because it directly predicts the sequence based on the low-resolution high-level features without RoI operations. Quantitatively, if the texts with an area smaller than 3000 (after resizing) are ignored during evaluation, the F-measure with generic lexicons on ICDAR 2015 will be improved to 77.5. Furthermore, current state-of-the-art methods on ICDAR 2015 usually adopt larger image sizes during training and testing. For example, the short sides of the testing images are resized to 1440 pixels, while the long sides are shorter than 4000 pixels. As shown in Tab.~\ref{ICDAR 2015 End-to-End recognition result v2}, the performance of SPTSv2 on ICDAR 2015 with a larger testing size is much better than that with a smaller testing size.
    
    \begin{table}[htb!]
        \centering
        \caption{Ablation study for different testing scales on ICDAR 2015. “S”, “W”, and “G” represent recognition with “Strong”, “Weak”, and “Generic” lexicon, respectively.}
        \label{ICDAR 2015 End-to-End recognition result v2}
        \small
        \begin{tabular}{r|c|c|c}
        \hline
        \multirow{2}{*}{Method} & \multicolumn{3}{c}{IC15 End-to-End}                    \\ \cline{2-4} 
                                & \multicolumn{1}{c|}{S}    & \multicolumn{1}{c|}{W}    & G    \\ \hline
        \methodName\ (1000) &  77.5 & 70.2 & 65.8 \\
        \methodName\ (1440) &  79.5 & 74.1 & 70.2 \\ 
        \hline
        SPTS v2\ (720) & 73.2 & 65.0 & 57.3 \\
        SPTS v2\ (1000) & 82.3 & 75.5 & 70.2 \\
        SPTS v2\ (1440)  &  82.3 & 77.7 & 72.6 \\ 
        \hline
        \end{tabular}
    \end{table}
    
    \subsubsection{Arbitrarily Shaped Dataset}
    We further compare our method with existing approaches on the benchmarks containing arbitrarily shaped texts, including Total-Text~\cite{ch2017total} and SCUT-CTW1500~\cite{liu2019curved}. As shown in Tab.~\ref{tab:cp_sota_e2e}, for single-point-based methods, our method achieves state-of-the-art performance, outperforming TOSS by a large margin. Additionally, Tab.~\ref{tab:cp_sota_e2e} shows that our method achieves superior results on the long text-line-based SCUT-CTW1500 dataset, which further demonstrates that the single-point could be strong enough to guide the text spotting. 
    For the challenging Inverse-Text, our method further achieves superior performance under the same setting as the previous method~\cite{huang2022swintextspotter}, with 8.0\% and 7.0\% higher than previous state-of-the-art in terms of the ``None'' and ``Full'' metrics, respectively, demonstrating its robustness to deal with rotated arbitrarily-shaped text.

    \begin{table*}[h!]
        \centering
        \caption{Comparison between the end-to-end recognition results of the SPTS and NPTS models. Loc$_{synth}$ denotes the use of location supervision exclusively for synthetic data, which requires no manual annotation. On the other hand, Loc$_{real}$ signifies the use of location supervision for real data. 
        }
        \label{SPTS-NoPoint}
        \resizebox{\linewidth}{!}{%
        \begin{tabular}{r|cc|cc|cc|ccc|ccc}
            \hline
            \multirow{2}*{Method} & \multirow{2}*{Loc$_{synth}$} & \multirow{2}*{Loc$_{real}$} & \multicolumn{2}{c|}{Total-Text} & \multicolumn{2}{c|}{SCUT-CTW1500} & \multicolumn{3}{c|}{ICDAR 2013} & \multicolumn{3}{c}{ICDAR 2015} \\
            \cline{4-13}
             & & & None & Full & None & Full & S & W & G & S & W & G \\
            \hline
            TOSS~\cite{tang2022you} &  & & 61.5 & 73.0 & 51.4 & 61.7 & 77.7 & 76.8 & 73.3 & 60.2 & 54.5 & 47.1 \\ 
            TTS~\cite{kittenplon2022towards} & \checkmark & & 75.1 & 83.5 & - & - & - & - & - & 78.7 & 75.2 & 70.1 \\ 
            SPTS \cite{peng2022spts} &  \checkmark & \checkmark & 74.2 & 82.4 & 63.6 & 83.8 & 93.3 & 91.7 & 88.5 & 77.5 & 70.2 & 65.8 \\
            \hline 
            SPTS v2 & \checkmark & \checkmark & \textbf{75.5} & \textbf{84.0} & \textbf{63.6} & \textbf{84.3} & \textbf{93.9} & \textbf{91.8} & \textbf{88.6} & \textbf{82.3} & \textbf{77.7} & \textbf{72.6} \\ 
            NPTS &   & & 64.7 & 71.9 & 55.4 & 74.3 & 90.4 & 84.9 & 80.2 & 69.4 & 60.3 & 55.6 \\
            \hline
        \end{tabular}}
    \end{table*}

    \begin{figure}[t!]
        \centering
        \begin{minipage}{0.9\linewidth}
            \includegraphics[width=8.3cm, height=4.5cm]{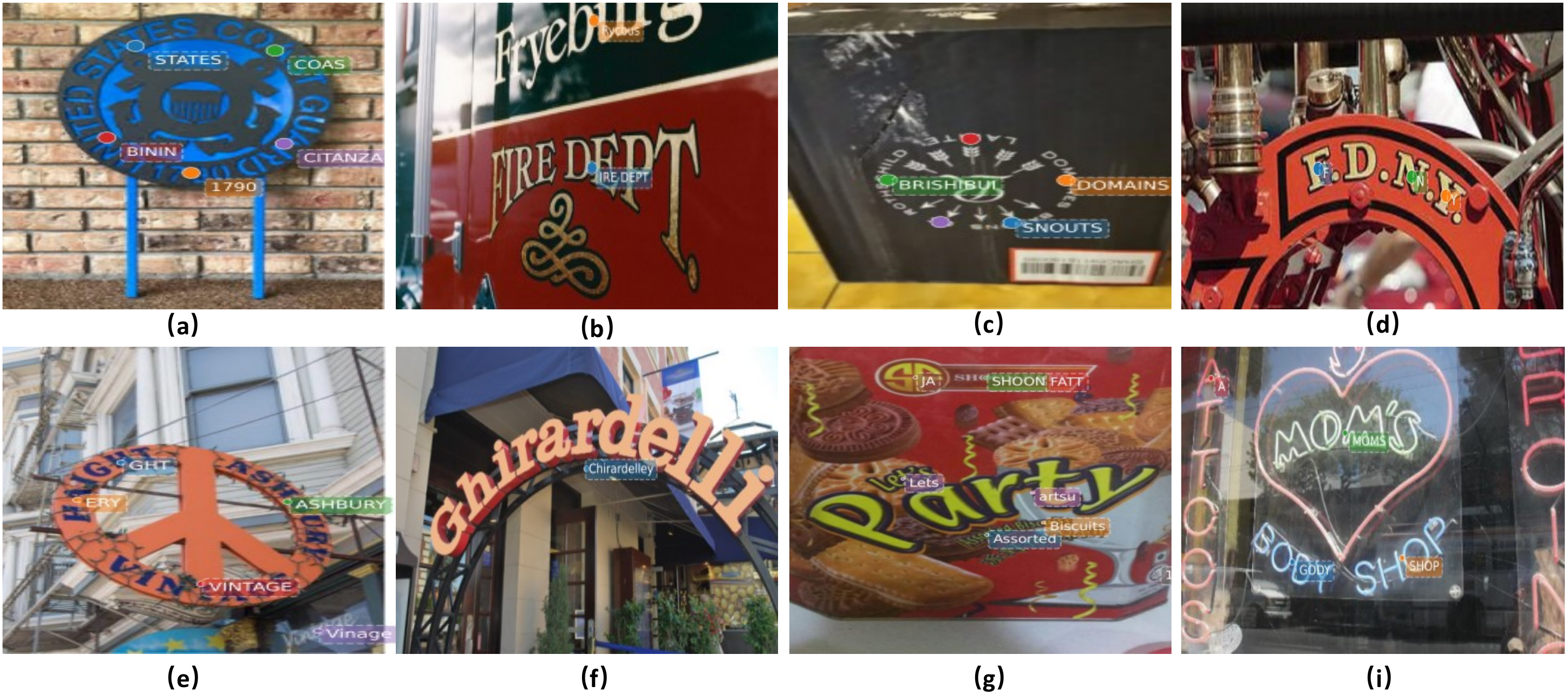}
        \end{minipage}
        \caption{Visualization of some of the failure cases of our method. Best view on screen.}
        \label{fig:failure_cases}
    \end{figure}

    \subsubsection{Summary}        
    In summary, the proposed method can achieve competitive performance compared with previous text spotters on several benchmarks. Especially on the two curved datasets, \emph{i.e.}, SCUT-CTW1500~\cite{liu2019curved}, the proposed method outperforms some recently proposed methods by a large margin. The reason why our methods can achieve better accuracy on arbitrary-shaped texts might be: (1) The proposed method discards the task-specific modules (\emph{e.g.}, RoI modules) designed based on prior knowledge; therefore, the recognition accuracy is decoupled with the detection results, \emph{i.e.}, our method can achieve acceptable recognition results even the detection position is shifted. 
    On the other hand, the features fed to the recognition module are sampled based on the ground-truth position during training but from detection results during testing, which leads to feature misalignment. However, by tackling the spotting task in a sequence modeling manner, the proposed method eliminates such issues, thus showing more robustness on especially long text-line based arbitrarily shaped datasets.
    
    Some of the visualization results of five datasets are shown in Fig.~\ref{fig:qual}. From the figure, we can see that the method shows robustness in curved, dense, highly-rotated, and long text. In the rightmost image of the second row, the multi-oriented dense long text can cause overlap, leading to missed instances in bounding-box based methods. However, with our approach, which uses a single point for location indication, such interference is naturally reduced, enabling accurate spotting of most instances.

    \begin{figure}[t!]
        \centering
        \includegraphics[width=0.95\linewidth]{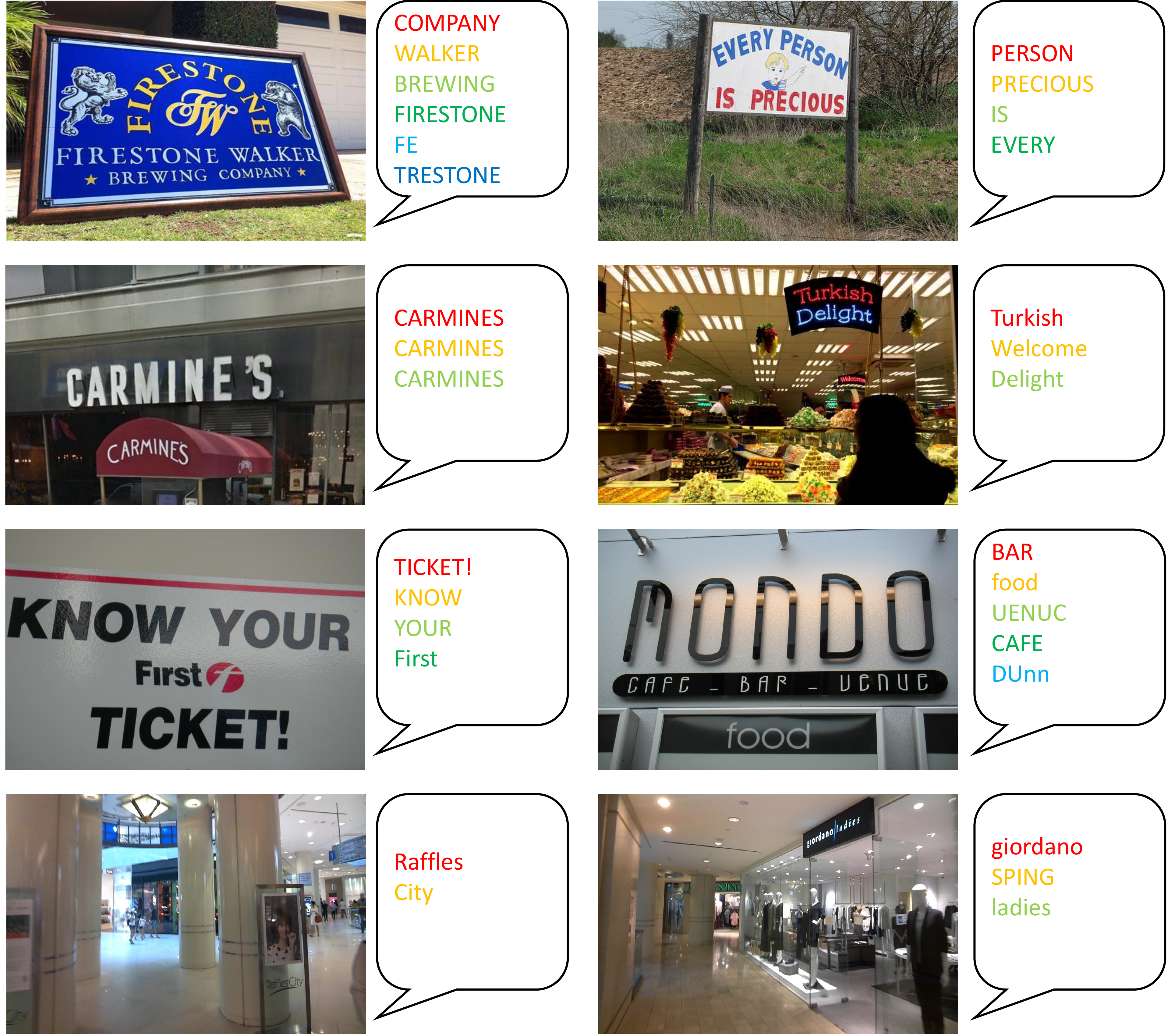}
        \caption{Qualitative results of the NPTS model on several scene text benchmarks. Images are selected from Total-Text (first row), SCUT-CTW1500 (second row), ICDAR 2013 (third row), and ICDAR 2015 (fourth row). Best viewed on screen.
        }
        \label{fig:vis_wopoint}
    \end{figure}

    \section{Discussion}
    We further conduct experiments to comprehensively evaluate the limitations and other property of our method.
    
    \subsection{Transcription-only Text Spotting}
    \label{subsec:non-point}
    As was prompted by TTS$_{weak}$~\cite{kittenplon2022towards}, a method that shows promising results in transcription-only fine-tuning with just synthetic data pretraining. We further establish that our approach can also converge without the supervision of location annotations. We develop a No-Point Text Spotting (NPTS) model by removing the coordinates of indicated points from the constructed sequence in both synthetic and real data. A comparison of our models with other methods is outlined in Tab.~\ref{SPTS-NoPoint}. As for evaluation metric, we substitute the distance matrix between predicted and ground truth (GT) points with an edit distance matrix between predicted and GT transcriptions. Other aspects are consistent with those outlined in Sec. \ref{subsec:eval_protocol}. When we employ location information from real data, SPTS v2 delivers the best performance. In a transcription-only context, TTS outperforms both TOSS and the NPTS model. This might be attributed to the valuable supervision provided by location information in freely available synthetic data, whereas TOSS and NPTS rely solely on synthetic data transcriptions. This suggests the significance of location indication for the text spotting task. In scenarios where location information from synthetic data is not used, NPTS performs better than TOSS. Fig. \ref{fig:vis_wopoint} displays qualitative results of NPTS, suggesting that the model has acquired the capacity to implicitly locate text based solely on transcriptions.

    \subsection{Failure Cases}
    We conducted a qualitative error analysis on the failure results. Fig.~\ref{fig:failure_cases} presents the visualization of some representative errors. In the case of Fig.~\ref{fig:failure_cases}(b), the error arises due to severe perspective distortion in the text and interference caused by illumination. In Fig.~\ref{fig:failure_cases}(a), an error occurs in the recognition of certain rotated characters. For instance, the character ``U'' is mistakenly identified as an ``I'' in the rightmost text. In Fig.~\ref{fig:failure_cases}(c), our method successfully detects the inverse text at the top; however, it fails to generate any recognition result for this particular case. For the case in Fig.~\ref{fig:failure_cases}(d), the errors occur due to the presence of full stop symbols that separate the characters. Furthermore, in Fig.~\ref{fig:failure_cases}(e), our method separates one text into two, resulting in an incorrect recognition prediction. Although recognition is not constrained by text boundaries, the incorrect distinction between different text instances ultimately leads to the failure in recognizing the text. In the case of Fig.~\ref{fig:failure_cases}(f), the model predicts recognition results that are completely inconsistent with the visual data, which may be inferred based on the previous predictions. Additionally, the proposed method still struggles with some artistic words, as shown in Fig.~\ref{fig:failure_cases}(g). For horizontally arranged vertical texts, our method fails to recognize them due to the lack of such training samples, as presented in Fig.~\ref{fig:failure_cases}(i).

    \section{Conclusion}
    \label{sec:conclude}
    We have proposed SPTS v2, a new scene text spotting paradigm that shows an extremely low-cost single-point annotation can be successfully used to train a powerful text spotter. SPTS v2 is based on a concise Transformer-based framework, in which the detection and recognition of the text are simply formulated as language sequences, requiring only the cross-entropy loss without feature alignment nor additional post-processing strategies. It includes an instance assignment decoder (IAD) which reserves the advantage of unifying all text instances inside the identical sequence, and a parallel recognition decoder (PRD) as well as the simple but effective information transmission method for significantly reducing the length of the sequence. Note both the IAD and PRD share exact the same parameters. With less parameters, SPTS v2 outperforms previous state-of-the-art single-point text spotter (SPTS) meanwhile with 19$\times$  faster for the inference speed. Extensive experiments demonstrate that such point-based method can still achieve competitive results. A key advantage of SPTS v2 is its ability to greatly reduce the required sequence length. The straightforward nature of its training approach makes it especially suited for multi-task scenarios, enabling diverse multimodal data to be processed through a cohesive token-to-language pipeline. Investigating this using larger models could offer a promising avenue for exploration.
    
    {
    \bibliographystyle{ieeetr}
     \bibliography{CSRef}

\begin{thebibliography}{10}

\bibitem{ch2017total}
C.~K. Ch'ng and C.~S. Chan, ``{Total-Text}: A comprehensive dataset for scene
  text detection and recognition,'' in {\em Proc. IAPR Int.\ Conf.\ Document
  Analysis Recog.}, vol.~1, pp.~935--942, IEEE, 2017.

\bibitem{liu2017deep}
Y.~Liu and L.~Jin, ``{Deep matching prior network: Toward tighter
  multi-oriented text detection},'' in {\em Proc. IEEE Conf. Comp. Vis. Patt.
  Recogn.}, pp.~3454--3461, 2017.

\bibitem{tian2016detecting}
Z.~Tian, W.~Huang, T.~He, P.~He, and Y.~Qiao, ``{Detecting text in natural
  image with connectionist text proposal network},'' in {\em Proc. Eur. Conf.
  Comp. Vis.}, pp.~56--72, Springer, 2016.

\bibitem{liao2017textboxes}
M.~Liao, B.~Shi, X.~Bai, X.~Wang, and W.~Liu, ``{TextBoxes: A fast text
  detector with a single deep neural network},'' in {\em Proc. {AAAI} Conf.
  Artificial Intell.}, pp.~4161--4167, 2017.

\bibitem{zhou2017east}
X.~Zhou, C.~Yao, H.~Wen, Y.~Wang, S.~Zhou, W.~He, and J.~Liang, ``{EAST}: An
  efficient and accurate scene text detector,'' in {\em Proc. IEEE Conf. Comp.
  Vis. Patt. Recogn.}, pp.~5551--5560, 2017.

\bibitem{yao2012detecting}
C.~Yao, X.~Bai, W.~Liu, Y.~Ma, and Z.~Tu, ``Detecting texts of arbitrary
  orientations in natural images,'' in {\em Proc. IEEE Conf. Comp. Vis. Patt.
  Recogn.}, pp.~1083--1090, 2012.

\bibitem{liu2020abcnet}
Y.~Liu, H.~Chen, C.~Shen, T.~He, L.~Jin, and L.~Wang, ``{ABCNet}: Real-time
  scene text spotting with adaptive bezier-curve network,'' in {\em Proc. IEEE
  Conf. Comp. Vis. Patt. Recogn.}, pp.~9809--9818, 2020.

\bibitem{lyu2018mask}
P.~Lyu, M.~Liao, C.~Yao, W.~Wu, and X.~Bai, ``{Mask TextSpotter}: An end-to-end
  trainable neural network for spotting text with arbitrary shapes,'' in {\em
  Proc. Eur. Conf. Comp. Vis.}, pp.~67--83, 2018.

\bibitem{shi2017detecting}
B.~Shi, X.~Bai, and S.~Belongie, ``Detecting oriented text in natural images by
  linking segments,'' in {\em Proc. IEEE Conf. Comp. Vis. Patt. Recogn.},
  pp.~2550--2558, 2017.

\bibitem{feng2019textdragon}
W.~Feng, W.~He, F.~Yin, X.-Y. Zhang, and C.-L. Liu, ``{TextDragon}: An
  end-to-end framework for arbitrary shaped text spotting,'' in {\em Proc. IEEE
  Int. Conf. Comp. Vis.}, pp.~9076--9085, 2019.

\bibitem{wang2020textray}
F.~Wang, Y.~Chen, F.~Wu, and X.~Li, ``{TextRay}: Contour-based geometric
  modeling for arbitrary-shaped scene text detection,'' in {\em Proc. {ACM}
  Int. Conf. Multimedia}, pp.~111--119, 2020.

\bibitem{zhu2021fourier}
Y.~Zhu, J.~Chen, L.~Liang, Z.~Kuang, L.~Jin, and W.~Zhang, ``Fourier contour
  embedding for arbitrary-shaped text detection,'' in {\em Proc. IEEE Conf.
  Comp. Vis. Patt. Recogn.}, pp.~3123--3131, 2021.

\bibitem{long2018textsnake}
S.~Long, J.~Ruan, W.~Zhang, X.~He, W.~Wu, and C.~Yao, ``{TextSnake: A flexible
  representation for detecting text of arbitrary shapes},'' in {\em Proc. Eur.
  Conf. Comp. Vis.}, pp.~20--36, 2018.

\bibitem{liao2019mask}
M.~Liao, P.~Lyu, M.~He, C.~Yao, W.~Wu, and X.~Bai, ``{Mask TextSpotter: An
  end-to-end trainable neural network for spotting text with arbitrary
  shapes},'' {\em {IEEE} Trans. Pattern Anal. Mach. Intell.}, pp.~532--548,
  2019.

\bibitem{tian2017wetext}
S.~Tian, S.~Lu, and C.~Li, ``{WeText}: Scene text detection under weak
  supervision,'' in {\em Proc. IEEE Int. Conf. Comp. Vis.}, pp.~1492--1500,
  2017.

\bibitem{bartz2018see}
C.~Bartz, H.~Yang, and C.~Meinel, ``{SEE}: Towards semi-supervised end-to-end
  scene text recognition,'' in {\em Proc. {AAAI} Conf. Artificial Intell.},
  2018.

\bibitem{hu2017wordsup}
H.~Hu, C.~Zhang, Y.~Luo, Y.~Wang, J.~Han, and E.~Ding, ``{WordSup}: Exploiting
  word annotations for character based text detection,'' in {\em Proc. IEEE
  Int. Conf. Comp. Vis.}, pp.~4940--4949, 2017.

\bibitem{baek2019character}
Y.~Baek, B.~Lee, D.~Han, S.~Yun, and H.~Lee, ``Character region awareness for
  text detection,'' in {\em Proc. IEEE Conf. Comp. Vis. Patt. Recogn.},
  pp.~9365--9374, 2019.

\bibitem{peng2022spts}
D.~Peng, X.~Wang, Y.~Liu, J.~Zhang, M.~Huang, S.~Lai, J.~Li, S.~Zhu, D.~Lin,
  C.~Shen, {\em et~al.}, ``{SPTS}: Single-point text spotting,'' in {\em Proc.
  {ACM} Int. Conf. Multimedia}, pp.~4272--4281, 2022.

\bibitem{li2017towards}
H.~Li, P.~Wang, and C.~Shen, ``{Towards end-to-end text spotting with
  convolutional recurrent neural networks},'' in {\em Proc. IEEE Int. Conf.
  Comp. Vis.}, pp.~5238--5246, 2017.

\bibitem{liao2020masktext}
M.~Liao, G.~Pang, J.~Huang, T.~Hassner, and X.~Bai, ``{Mask TextSpotter v3}:
  Segmentation proposal network for robust scene text spotting,'' in {\em Proc.
  Eur. Conf. Comp. Vis.}, pp.~706--722, 2020.

\bibitem{chen2021pix2seq}
T.~Chen, S.~Saxena, L.~Li, D.~J. Fleet, and G.~Hinton, ``{Pix2Seq}: A language
  modeling framework for object detection,'' in {\em Proc. Int. Conf. Learn.
  Representations}, 2022.

\bibitem{carion2020end}
N.~Carion, F.~Massa, G.~Synnaeve, N.~Usunier, A.~Kirillov, and S.~Zagoruyko,
  ``End-to-end object detection with transformers,'' in {\em Proc. Eur. Conf.
  Comp. Vis.}, pp.~213--229, 2020.

\bibitem{neubeck2006efficient}
A.~Neubeck and L.~Van~Gool, ``Efficient non-maximum suppression,'' in {\em 18th
  International Conference on Pattern Recognition}, vol.~3, pp.~850--855, 2006.

\bibitem{karatzas2013icdar}
D.~Karatzas, F.~Shafait, S.~Uchida, {\em et~al.}, ``{ICDAR 2013 robust reading
  competition},'' in {\em Proc. IAPR Int.\ Conf.\ Document Analysis Recog.},
  pp.~1484--1493, 2013.

\bibitem{karatzas2015icdar}
D.~Karatzas, L.~Gomez-Bigorda, {\em et~al.}, ``{ICDAR 2015 competition on
  robust reading},'' in {\em Proc. IAPR Int.\ Conf.\ Document Analysis Recog.},
  pp.~1156--1160, 2015.

\bibitem{liu2019curved}
Y.~Liu, L.~Jin, S.~Zhang, C.~Luo, and S.~Zhang, ``{Curved scene text detection
  via transverse and longitudinal sequence connection},'' {\em Pattern
  Recognition}, vol.~90, pp.~337--345, 2019.

\bibitem{ye2022dptext}
M.~Ye, J.~Zhang, S.~Zhao, J.~Liu, B.~Du, and D.~Tao, ``{DPText-DETR}: Towards
  better scene text detection with dynamic points in transformer,'' in {\em
  Proc. {AAAI} Conf. Artificial Intell.}, 2023.

\bibitem{xing2019convolutional}
L.~Xing, Z.~Tian, W.~Huang, and S.~M. R., ``{Convolutional character
  networks},'' in {\em Proc. IEEE Int. Conf. Comp. Vis.}, pp.~9126--9136, 2019.

\bibitem{qiao2021mango}
L.~Qiao, Y.~Chen, Z.~Cheng, Y.~Xu, Y.~Niu, S.~Pu, and F.~Wu, ``{MANGO}: A mask
  attention guided one-stage scene text spotter,'' in {\em Proc. {AAAI} Conf.
  Artificial Intell.}, pp.~2467--2476, 2021.

\bibitem{karatzas2011icdar}
D.~Karatzas, S.~R. Mestre, J.~Mas, F.~Nourbakhsh, and P.~P. Roy, ``{ICDAR} 2011
  robust reading competition-challenge 1: reading text in born-digital images
  (web and email),'' in {\em Proc. IAPR Int.\ Conf.\ Document Analysis Recog.},
  pp.~1485--1490, 2011.

\bibitem{nayef2019icdar2019}
N.~Nayef, Y.~Patel, M.~Busta, P.~N. Chowdhury, D.~Karatzas, W.~Khlif, J.~Matas,
  U.~Pal, J.-C. Burie, C.-l. Liu, {\em et~al.}, ``{ICDAR2019} robust reading
  challenge on multi-lingual scene text detection and
  recognition--{RRC-MLT}-2019,'' {\em Proc. IAPR Int.\ Conf.\ Document Analysis
  Recog.}, pp.~1582--1587, 2019.

\bibitem{chng2019icdar2019}
C.-K. Chng, Y.~Liu, Y.~Sun, C.~C. Ng, C.~Luo, Z.~Ni, C.~Fang, S.~Zhang, J.~Han,
  E.~Ding, {\em et~al.}, ``{ICDAR2019} robust reading challenge on
  arbitrary-shaped text ({RRC-ArT}),'' {\em Proc. IAPR Int.\ Conf.\ Document
  Analysis Recog.}, pp.~1571--1576, 2019.

\bibitem{fan2022pointly}
J.~Fan, Z.~Zhang, and T.~Tan, ``Pointly-supervised panoptic segmentation,'' in
  {\em Proc. Eur. Conf. Comp. Vis.}, pp.~319--336, Springer, 2022.

\bibitem{tang2022you}
J.~Tang, S.~Qiao, B.~Cui, Y.~Ma, S.~Zhang, and D.~Kanoulas, ``You can even
  annotate text with voice: Transcription-only-supervised text spotting,'' in
  {\em Proc. {ACM} Int. Conf. Multimedia}, pp.~4154--4163, 2022.

\bibitem{wang2010word}
K.~Wang and S.~Belongie, ``Word spotting in the wild,'' in {\em Proc. Eur.
  Conf. Comp. Vis.}, pp.~591--604, 2010.

\bibitem{dalal2005histograms}
N.~Dalal and B.~Triggs, ``Histograms of oriented gradients for human
  detection,'' in {\em Proc. IEEE Conf. Comp. Vis. Patt. Recogn.}, vol.~1,
  pp.~886--893, 2005.

\bibitem{bissacco2013photoocr}
A.~Bissacco, M.~Cummins, Y.~Netzer, and H.~Neven, ``{PhotoOCR}: Reading text in
  uncontrolled conditions,'' in {\em Proc. IEEE Int. Conf. Computer Vision},
  pp.~785--792, 2013.

\bibitem{jaderberg2014deep}
M.~Jaderberg, A.~Vedaldi, and A.~Zisserman, ``Deep features for text
  spotting,'' in {\em Proc. Eur. Conf. Comp. Vis.}, pp.~512--528, 2014.

\bibitem{baek2020character}
Y.~Baek, S.~Shin, J.~Baek, S.~Park, J.~Lee, D.~Nam, and H.~Lee, ``Character
  region attention for text spotting,'' in {\em Proc. Eur. Conf. Comp. Vis.},
  pp.~504--521, 2020.

\bibitem{weinman2013toward}
J.~J. Weinman, Z.~Butler, D.~Knoll, and J.~Feild, ``Toward integrated scene
  text reading,'' {\em {IEEE} Trans. Pattern Anal. Mach. Intell.}, vol.~36,
  no.~2, pp.~375--387, 2013.

\bibitem{ren2015faster}
S.~Ren, K.~He, R.~Girshick, and J.~Sun, ``{Faster R-CNN}: Towards real-time
  object detection with region proposal networks,'' in {\em Proc. Advances in
  Neural Inf. Process. Syst.}, pp.~91--99, 2015.

\bibitem{graves2006connectionist}
A.~Graves, S.~Fern{\'a}ndez, F.~Gomez, and J.~Schmidhuber, ``{Connectionist
  temporal classification: Labelling unsegmented sequence data with recurrent
  neural networks},'' in {\em Proc. Int. Conf. Mach. Learn.}, pp.~369--376,
  ACM, 2006.

\bibitem{wang2021towards}
P.~Wang, H.~Li, and C.~Shen, ``Towards end-to-end text spotting in natural
  scenes,'' {\em {IEEE} Trans. Pattern Anal. Mach. Intell.}, vol.~44, no.~10,
  pp.~7266--7281, 2022.

\bibitem{gupta2016synthetic}
A.~Gupta, A.~Vedaldi, and A.~Zisserman, ``Synthetic data for text localisation
  in natural images,'' in {\em Proc. IEEE Conf. Comp. Vis. Patt. Recogn.},
  pp.~2315--2324, 2016.

\bibitem{liao2018textboxes++}
M.~Liao, B.~Shi, and X.~Bai, ``{TextBoxes++}: A single-shot oriented scene text
  detector,'' {\em {IEEE} Trans. Image Process.}, vol.~27, no.~8,
  pp.~3676--3690, 2018.

\bibitem{shi2017end}
B.~Shi, X.~Bai, and C.~Yao, ``An end-to-end trainable neural network for
  image-based sequence recognition and its application to scene text
  recognition,'' {\em IEEE Trans. Pattern Anal. Mach. Intell}, vol.~39, no.~11,
  pp.~2298--2304, 2017.

\bibitem{shi2018aster}
B.~Shi, M.~Yang, X.~Wang, P.~Lyu, C.~Yao, and X.~Bai, ``{ASTER: An attentional
  scene text recognizer with flexible rectification},'' {\em {IEEE} Trans.
  Pattern Anal. Mach. Intell.}, vol.~41, no.~9, pp.~2035--2048, 2019.

\bibitem{bookstein1989principal}
F.~L. Bookstein, ``{Principal warps: Thin-plate splines and the decomposition
  of deformations},'' {\em {IEEE} Trans. Pattern Anal. Mach. Intell.}, vol.~11,
  no.~6, pp.~567--585, 1989.

\bibitem{busta2017deep}
M.~Busta, L.~Neumann, and J.~Matas, ``{Deep TextSpotter}: An end-to-end
  trainable scene text localization and recognition framework,'' in {\em Proc.
  IEEE Int. Conf. Comp. Vis.}, pp.~2204--2212, 2017.

\bibitem{redmon2017yolo9000}
J.~Redmon and A.~Farhadi, ``{YOLO9000}: Better, faster, stronger,'' in {\em
  Proc. IEEE Conf. Comp. Vis. Patt. Recogn.}, pp.~7263--7271, 2017.

\bibitem{liu2018fots}
X.~Liu, D.~Liang, S.~Yan, D.~Chen, Y.~Qiao, and J.~Yan, ``{FOTS}: Fast oriented
  text spotting with a unified network,'' in {\em Proc. IEEE Conf. Comp. Vis.
  Patt. Recogn.}, pp.~5676--5685, 2018.

\bibitem{he2018end}
T.~He, Z.~Tian, W.~Huang, C.~Shen, Y.~Qiao, and C.~Sun, ``An end-to-end
  textspotter with explicit alignment and attention,'' in {\em Proc. IEEE Conf.
  Comp. Vis. Patt. Recogn.}, pp.~5020--5029, 2018.

\bibitem{qin2019towards}
S.~Qin, A.~Bissacco, M.~Raptis, Y.~Fujii, and Y.~Xiao, ``Towards unconstrained
  end-to-end text spotting,'' {\em Proc. IEEE Int. Conf. Comp. Vis.},
  pp.~4704--4714, 2019.

\bibitem{wang2021pan++}
W.~Wang, E.~Xie, X.~Li, X.~Liu, D.~Liang, Y.~Zhibo, T.~Lu, and C.~Shen,
  ``{PAN++}: Towards efficient and accurate end-to-end spotting of
  arbitrarily-shaped text,'' {\em {IEEE} Trans. Pattern Anal. Mach. Intell.},
  vol.~44, no.~9, pp.~5349--5367, 2022.

\bibitem{wang2019efficient}
W.~Wang, E.~Xie, X.~Song, Y.~Zang, W.~Wang, T.~Lu, G.~Yu, and C.~Shen,
  ``Efficient and accurate arbitrary-shaped text detection with pixel
  aggregation network,'' {\em Proc. IEEE Int. Conf. Comp. Vis.},
  pp.~8440--8449, 2019.

\bibitem{sun2018textnet}
Y.~Sun, C.~Zhang, Z.~Huang, J.~Liu, J.~Han, and E.~Ding, ``{TextNet}: Irregular
  text reading from images with an end-to-end trainable network,'' in {\em
  Proc. Asian Conf. Comp. Vis.}, pp.~83--99, Springer, 2018.

\bibitem{wang2020all}
H.~Wang, P.~Lu, H.~Zhang, M.~Yang, X.~Bai, Y.~Xu, M.~He, Y.~Wang, and W.~Liu,
  ``All you need is boundary: Toward arbitrary-shaped text spotting,'' in {\em
  Proc. {AAAI} Conf. Artificial Intell.}, vol.~34, pp.~12160--12167, 2020.

\bibitem{qiao2020text}
L.~Qiao, S.~Tang, Z.~Cheng, Y.~Xu, Y.~Niu, S.~Pu, and F.~Wu, ``Text perceptron:
  Towards end-to-end arbitrary-shaped text spotting,'' in {\em Proc. {AAAI}
  Conf. Artificial Intell.}, vol.~34, pp.~11899--11907, 2020.

\bibitem{tian2019fcos}
Z.~Tian, C.~Shen, H.~Chen, and T.~He, ``{FCOS}: Fully convolutional one-stage
  object detection,'' in {\em Proc. IEEE Int. Conf. Comp. Vis.},
  pp.~9627--9636, 2019.

\bibitem{liu2021abcnetv2}
Y.~Liu, C.~Shen, L.~Jin, T.~He, P.~Chen, C.~Liu, and H.~Chen, ``{ABCNet} v2:
  Adaptive bezier-curve network for real-time end-to-end text spotting,'' {\em
  {IEEE} Trans. Pattern Anal. Mach. Intell.}, vol.~44, no.~11, pp.~8048--8064,
  2022.

\bibitem{tan2020efficientdet}
M.~Tan, R.~Pang, and Q.~V. Le, ``{EfficientDet}: Scalable and efficient object
  detection,'' in {\em Proc. IEEE Conf. Comp. Vis. Patt. Recogn.},
  pp.~10781--10790, 2020.

\bibitem{huang2022swintextspotter}
M.~Huang, Y.~Liu, Z.~Peng, C.~Liu, D.~Lin, S.~Zhu, N.~Yuan, K.~Ding, and
  L.~Jin, ``{SwinTextSpotter}: Scene text spotting via better synergy between
  text detection and text recognition,'' in {\em Proc. IEEE Conf. Comp. Vis.
  Patt. Recogn.}, pp.~4593--4603, 2022.

\bibitem{zhang2022text}
X.~Zhang, Y.~Su, S.~Tripathi, and Z.~Tu, ``Text spotting transformers,'' in
  {\em Proc. IEEE Conf. Comp. Vis. Patt. Recogn.}, pp.~9519--9528, 2022.

\bibitem{zhu2020deformable}
X.~Zhu, W.~Su, L.~Lu, B.~Li, X.~Wang, and J.~Dai, ``Deformable {DETR}:
  Deformable transformers for end-to-end object detection,'' {\em Proc. Int.
  Conf. Learn. Representations}, 2021.

\bibitem{fang2022abinet++}
S.~Fang, Z.~Mao, H.~Xie, Y.~Wang, C.~Yan, and Y.~Zhang, ``{ABINet++}:
  Autonomous, bidirectional and iterative language modeling for scene text
  spotting,'' {\em {IEEE} Trans. Pattern Anal. Mach. Intell.}, 2022.

\bibitem{fang2021read}
S.~Fang, H.~Xie, Y.~Wang, Z.~Mao, and Y.~Zhang, ``Read like humans: Autonomous,
  bidirectional and iterative language modeling for scene text recognition,''
  in {\em Proc. IEEE Conf. Comp. Vis. Patt. Recogn.}, pp.~7098--7107, 2021.

\bibitem{kittenplon2022towards}
Y.~Kittenplon, I.~Lavi, S.~Fogel, Y.~Bar, R.~Manmatha, and P.~Perona, ``Towards
  weakly-supervised text spotting using a multi-task transformer,'' in {\em
  Proc. IEEE Conf. Comp. Vis. Patt. Recogn.}, pp.~4604--4613, 2022.

\bibitem{kuhn1955hungarian}
H.~W. Kuhn, ``The hungarian method for the assignment problem,'' {\em Naval
  research logistics quarterly}, vol.~2, no.~1-2, pp.~83--97, 1955.

\bibitem{meng2021-CondDETR}
D.~Meng, X.~Chen, Z.~Fan, G.~Zeng, H.~Li, Y.~Yuan, L.~Sun, and J.~Wang,
  ``Conditional {DETR} for fast training convergence,'' in {\em Proc.\ IEEE
  Int.\ Conf.\ Computer Vision}, pp.~3631--3640, 2021.

\bibitem{liu2022dabdetr}
S.~Liu, F.~Li, H.~Zhang, X.~Yang, X.~Qi, H.~Su, J.~Zhu, and L.~Zhang,
  ``{DAB}-{DETR}: Dynamic anchor boxes are better queries for {DETR},'' in {\em
  Int.\ Conf.\ Learning Representations}, 2022.

\bibitem{nayef2017icdar2017}
N.~Nayef, F.~Yin, I.~Bizid, H.~Choi, Y.~Feng, D.~Karatzas, Z.~Luo, U.~Pal,
  C.~Rigaud, J.~Chazalon, {\em et~al.}, ``{ICDAR 2017} robust reading challenge
  on multi-lingual scene text detection and script
  identification-{RRC}-{MLT},'' in {\em Proc. IAPR Int.\ Conf.\ Document
  Analysis Recog.}, vol.~1, pp.~1454--1459, IEEE, 2017.

\bibitem{loshchilov2017decoupled}
I.~Loshchilov and F.~Hutter, ``Decoupled weight decay regularization,'' {\em
  Proc. Int. Conf. Learn. Representations}, 2018.

\bibitem{hoffer2020augment}
E.~Hoffer, T.~Ben-Nun, I.~Hubara, N.~Giladi, T.~Hoefler, and D.~Soudry,
  ``Augment your batch: Improving generalization through instance repetition,''
  in {\em Proc. IEEE Conf. Comp. Vis. Patt. Recogn.}, pp.~8129--8138, 2020.

\bibitem{xiong2020layer}
R.~Xiong, Y.~Yang, D.~He, K.~Zheng, S.~Zheng, C.~Xing, H.~Zhang, Y.~Lan,
  L.~Wang, and T.~Liu, ``On layer normalization in the {Transformer}
  architecture,'' in {\em Proc. Int. Conf. Mach. Learn.}, pp.~10524--10533,
  2020.

\bibitem{wang2021pgnet}
P.~Wang, C.~Zhang, F.~Qi, S.~Liu, X.~Zhang, P.~Lyu, J.~Han, J.~Liu, E.~Ding,
  and G.~Shi, ``{PGNet}: Real-time arbitrarily-shaped text spotting with point
  gathering network,'' in {\em Proc. {AAAI} Conf. Artificial Intell.},
  pp.~2782--2790, 2021.

\bibitem{ronen2022glass}
R.~Ronen, S.~Tsiper, O.~Anschel, I.~Lavi, A.~Markovitz, and R.~Manmatha,
  ``{GLASS}: Global to local attention for scene-text spotting,'' in {\em Proc.
  Eur. Conf. Comp. Vis.}, pp.~249--266, Springer, 2022.

\bibitem{luboundary}
P.~Lu, H.~Wang, S.~Zhu, J.~Wang, X.~Bai, and W.~Liu, ``Boundary {TextSpotter}:
  Toward arbitrary-shaped scene text spotting,'' {\em IEEE Transactions on
  Image Processing}, vol.~31, pp.~6200--6212, 2022.

\bibitem{wu2022decoupling}
J.~Wu, P.~Lyu, G.~Lu, C.~Zhang, K.~Yao, and W.~Pei, ``Decoupling recognition
  from detection: Single shot self-reliant scene text spotter,'' in {\em Proc.
  {ACM} Int. Conf. Multimedia}, pp.~1319--1328, 2022.

\bibitem{jaderberg2016reading}
M.~Jaderberg, K.~Simonyan, A.~Vedaldi, and A.~Zisserman, ``Reading text in the
  wild with convolutional neural networks,'' {\em Int. J. Comput. Vision},
  vol.~116, no.~1, pp.~1--20, 2016.

\end{thebibliography}
    }
    \end{document}